\documentclass[information,article,accept,moreauthors,pdftex]{Definitions/mdpi} 

\usepackage{booktabs} 
\usepackage{multirow}
\usepackage{soul} 
\usepackage{microtype}

\usepackage{mdframed}
\usepackage[ruled, linesnumbered]{algorithm2e}
\SetKwInput{KwParameters}{Parameters}

\firstpage{1} 
\pubvolume{12}
\issuenum{6}
\articlenumber{221}
\pubyear{2021}
\copyrightyear{2021}
\externaleditor{Academic Editor: Rodrigo Agerri} % For journal Automation, please change Academic Editor to "Communicated by"
\datereceived{3 April 2021} 
\dateaccepted{20 May 2021} 
\datepublished{23 May 2021} 
\hreflink{https://\\doi.org/10.3390/info12060221} % If needed use \linebreak
\Title{Term-Community-Based Topic Detection with Variable Resolution}

\TitleCitation{Term-Community-Based Topic Detection with Variable Resolution}

\Author{{Andreas Hamm} %MDPI: Please carefully check the accuracy of names and affiliations.%AH: Names and affiliations are correct.
 *\orcidA{} and Simon Odrowski \orcidB{}}

\AuthorNames{Andreas Hamm and Simon Odrowski}

\AuthorCitation{Hamm, A.; Odrowski, S.}
\address[1]{Think Tank, German Aerospace Center (DLR), {51147} Cologne, %MDPI: Please add post code. %AH: Post code has been added.
 Germany; simon.odrowski@dlr.de}
\corres{\hangafter=1 \hangindent=1.05em \hspace{-0.82em} Correspondence: andreas.hamm@dlr.de}

\abstract{Network-based procedures for topic detection in huge text collections offer an intuitive alternative to probabilistic topic models. We present in detail a method that is especially designed with the requirements of domain experts in mind. Like similar methods, it employs community detection in term co-occurrence graphs, but it is enhanced by including a resolution parameter that can be used for changing the targeted topic granularity. We also establish a term ranking and use semantic word-embedding for presenting term communities in a way that facilitates their interpretation.
We demonstrate the application of our method with a widely used corpus of general news articles and show the results of detailed social-sciences expert evaluations of detected topics at various resolutions. A comparison with topics detected by Latent Dirichlet Allocation is also included. Finally, we discuss factors that influence topic interpretation.}


\DeclareMathOperator{\pos}{pos}
\DeclareMathOperator{\Idf}{Idf}
\DeclareMathOperator{\Tf}{Tf}
\DeclareMathOperator{\Df}{Df}
\DeclareMathOperator{\fTsim}{fTsim}

\begin{document}
%%%%%%%%%%%%%%%%%%%%%%%%%%%%%%%%%%%%%%%%%%
%\setcounter{section}{-1} %% Remove this when starting to work on the template.
%The order of the section titles is: Introduction, Materials and Methods, Results, Discussion, Conclusions for these journals: aerospace,algorithms,antibodies,antioxidants,atmosphere,axioms,biomedicines,carbon,crystals,designs,diagnostics,environments,fermentation,fluids,forests,fractalfract,informatics,information,inventions,jfmk,jrfm,lubricants,neonatalscreening,neuroglia,particles,pharmaceutics,polymers,processes,technologies,viruses,vision

\section{Introduction}
Facing an ever-growing amount of text data, automated methods of text evaluation have become indispensable for finding and analyzing information. Computerized information retrieval started to evolve many decades ago as one of the earliest fields of computer applications but continues to make spectacular progress in the context of recent machine learning developments.

The classical information retrieval task is to serve some information need formulated as a concrete query. However, given the sheer volume of texts available, there are many situations where, before asking detailed questions, one must first gain some insight into what kind of information is contained in the texts at all and what subject areas are covered. 

This is where automatic topic detection (see references in Section \ref{sec:related_work}), also called topic mining or topic modeling, can help. This process takes a text corpus, i.e., a large collection of text documents, as input and produces as output a set of {topics} which are meant to represent the various subjects written about in the corpus documents. The identification of topics within a corpus can be used in many ways: for a quick overview of the content and a better understanding if little is known about the corpus or its context; for ordering the documents of the corpus, similar to a classification, but more flexible in that it allows one document to be assigned to several topics rather than belonging only to one class; or for analyzing the temporal evolution of thematic content or its relation to metadata like authorship or publisher. Computationally, it also can be seen as a method of dimensionality reduction for the corpus documents and, as such, lends itself as a building block in further machine learning analyses of the documents. 

Like with any computerized application, at both ends of the process some translation step is needed: at the input side a quantification which turns the corpus into some mathematical data structure, and at the output side an interpretation of what the algorithmically derived output actually means. The latter step can be highly problematic in situations involving natural language as it carries more ambiguity and context dependency than numerical or highly formalized data. This can be an obstacle for finding indisputable and verifiable interpretations. Therefore, involving subject experts who ideally are well trained in methods of text interpretation is crucial.

While this paper focuses on general, technical, and methodological aspects rather than on comprehensive domain applications, it is written as a collaboration of a computational data scientist and a political scientist in order to keep a good balance between the computational and the interpretive aspects. Political science is, in fact, one of the domains that benefit most from reliable methods for automated topic discovery: while text is an indispensable source of knowledge about politics, the discipline, in line with the general trend, has recently been confronted with a flood of relevant textual material \cite{Monroe2008, Brady2019, Benoit2020}. The background which motivated the present research is the need to scan and understand the strategic significance of huge amounts of incoming text documents of scientific, political, social, and  economic nature in a strategic unit of a large research organization.

Regarding the quantification of the corpus, the natural candidate for a data structure is the so-called word-document matrix that keeps record of which words of the total corpus are contained in which document and how important they are for the document. The earliest approaches to topic discovery, going back to the 1980s, applied purely algebraic considerations to that word-document matrix. The currently predominant approaches to topic discovery can be grouped into two distinctly different lines: one is based on probabilistic generative models where topics are parameters (more specifically: probability distributions on words) that can be determined by statistical inference. The other one is based on transforming the word-document matrix into a network in which the topics show up as communities of strongly linked nodes. We will mention references for some of the existing variants of both lines in the next section.

There is a striking imbalance between the popularity of the two lines. The number of research papers using probabilistic topic models exceeds the number of publications following network-based approaches of topic detection by two orders of magnitude. However, in spite of many impressive successful applications of the probabilistic models, it is not at all clear that they offer the best solutions in all situations. An initial investigation of several network-based methods in our group \cite{Thelen2020}, the findings of which we will sketch in Section \ref{sec:related_work}, showed a very promising potential and motivated further improvements of network-based topic detection which will be described in this paper.

The structure of the paper is as follows: in the next section we give a brief overview of some of the related work in the areas of topic mining and community detection. \mbox{Section \ref{sec:method}} presents our particular version of network-based topic detection. We exemplify the method by applying it to a well-known corpus of BBC news reports \cite{Greene2006} which has been widely used for text classification, topic modeling, and other text mining tasks in the literature (e.g., \cite{Anoop2016, Bodrunova2020}). In Section \ref{sec:influence}, we investigate the influence of two of the adjustable parameters of the method: the reduction percentage and the resolution parameter, and show how the latter one can be used to identify more and more topics on finer scale. For comparison, \mbox{in Section \ref{sec:interpretability_comparison}} we apply the best-known probabilistic topic modeling approach, Latent Dirichlet Allocation (LDA), to our example corpus and elaborate on observations regarding topic interpretability and other differences. Section \ref{seq:conclusions} draws some conclusions.

There are three main new contributions of this paper to the field of topic discovery: first, we describe a particular method for term ranking which is an essential ingredient for producing and interpreting high-quality topics. Second, we introduce for the first time in the context of term co-occurrence networks a topic detection method with which one can control the resulting topic granularity, which is particularly relevant from a domain expert perspective. We achieve this by using the Leiden algorithm for optimization of a generalized modularity, which is different from the community detection methods which have been previously used for topic identification. Third, we present new insight into questions of topic interpretability on the basis of expert evaluations and supported by \mbox{word embeddings. }

\section{Related Work}
\label{sec:related_work}
Methods of automatic topic detection (as well as other methods used in this article: keyword extraction and word embeddings) are based on the {distributional hypothesis} \cite{Harris1954}, which states that observations about the distribution of word occurrences allow to draw conclusions about semantics. The first systematic approach to topic detection was {Latent Semantic Indexing} (LSI) \cite{Deerwester1990}, based on singular value decomposition of the word-document matrix. Another successful algebraic method uses {non-negative matrix factorization} \mbox{(NMF) \cite{Xu2003}.}

{Going beyond purely algebraic operations, {probabilistic Latent Semantic Analysis} \mbox{(pLSA) \cite{Hofmann1999}}} regards the observed word distribution in the documents as the outcome of a stochastic process that results from the mixture of two multinomial distributions, which can be reconstructed using stochastic inference. LDA \cite{Blei2003} follows a similar strategy but goes one step further in assuming that the mixture is not between fixed but random multinomial distributions, which are drawn from a Dirichlet distribution. 

LDA has become enormously popular, not least because of several easy-to-use software implementations which employ efficient inference techniques like  {collapsed Gibbs sampling} \cite{Griffiths2004}.
It has been applied to many diverse text collections like scientific publications, news collections, literary corpora, political debates, historical documents, social media posts, and many others; for reviews we refer to \cite{Blei2012, BoydGraber2017, Jelodar2018}.
On the other hand, LDA motivated the development of a plethora of similar generative models with the aim of improving the method or of taking better account of special properties of the text collections to be studied. An example are generative models which can detect hierarchies of \mbox{topics \cite{Blei2003h, Grimmer2010}}.

Probabilistic topic models can be further enhanced by supplementing the word co-occurrence information with document metadata, like information on authorship, geographical location or relatedness to events \cite{Wang2012, Yan2012}. The Structural Topic Models (STM), which have proven useful in political science applications, also belong to that category \cite{Roberts2014}.

A fundamentally different line of topic detection methods arose from graph-theoreti\-cal evaluation of word-document co-occurrences; we refer to \cite{Sonawane2014} for a survey of graph-based text analysis. However, compared with probabilistic topic modeling, this line neither follows a homogeneous evolutionary history, nor has a widespread standard implementation been established yet.

Here we mention some early work connecting the concepts of topics and graphs: so-called co-word maps were produced in a semi-manual fashion in early bibliometric \mbox{studies \cite{Rip1984}}. TopCat \cite{Clifton1999} is one of the first graph-based procedures of topic detection. It is based on hypergraph clustering in co-occurrence hypergraphs of so called frequent-itemsets of named entities. Another approach is known under the name KeyGraph. It started as a method for key word extraction \cite{Ohsawa1998} based on a word co-occurrence graph on sentence level, but was later extended for event detection \cite{Ohsawa2003, Wang2013}.

While the early approaches were not suited for detailed analyses of large scale document collections, increased interest in network analysis and in particular the concept of community detection furthered the development of efficient graph-based topic discovery. We refer to \cite{Fortunato2016, Kumar2021} for reviews on community detection. Several methods of community detection have been used for topic discovery: Sayyadi and Raschid \cite{Sayyadi2013} find topics as communities in a KeyGraph by the Girvan--Newman algorithm involving the \mbox{edge-betweenness \cite{Girvan2002}}. Instead, Yang et. al. \cite{Yang2018} employ modularity maximization \cite{Newman2006}, using the Louvain algorithm \cite{Blondel2008}; similar approaches can be found in \cite{Salerno2015, Arruda2016, Dang2018, Kim2020}. Louvain-based community detection was also applied in \cite{Leydesdorff2016, Hecking2019}, in combination with a principle component analysis, to co-word maps. The Infomap algorithm \cite{Rosvall2009} for community detection via a random walk was used in \cite{Lancichinetti2015}. Wang et. al. \cite{Wang2017} identify topics as cliques in a word co-occurrence network. The hierarchical semantic graph model in \cite{Zhang2020} is based on a hierarchy of terms and uses subgraph segmentation via the normalized cut \mbox{algorithm \cite{Shi2000}} for community detection. Gerlach et. al. \cite{Gerlach2018} find topics as communities in a bipartite document-word graph with a Stochastic Block Model \cite{Karrer2011}. This approach establishes an interesting connection to the probabilistic topic models, as a Stochastic Block Model itself is a generative model. In fact, this graph-based topic detection method is closely related to pLSA. On the other hand, Stochastic Block Models also have been shown to be related to maximizing \cite{Newman2016} a parametrized generalized modularity~\cite{Reichardt2006}.

Several of the approaches listed in the previous paragraph appear to have evolved independently and largely unaware of each other. The potentially confusing diversity of varieties might even be one reason why---in spite of the case-by-case success of the community detection route to topic identification evident from the cited literature---most applied studies still consider only LDA or similar probabilistic methods for topic detection. \mbox{Thelen~\cite{Thelen2020}} compares, by the way of example, how the choice of community detection algorithm and network definition influences the results of graph-based topic detection. Specifically, she uses Louvain modularity optimization~\cite{Blondel2008} as well as the Infomap algorithm~\cite{Rosvall2009} and two different versions of edge weights for the co-occurrence networks, namely weighting by unadjusted co-occurrence counts like in~\cite{Salerno2015, Yang2018} or weighting by co-occurrence counts adjusted to a null model like in~\cite{Lancichinetti2015}. In addition, the Stochastic Block Model approach~\cite{Gerlach2018} is also included in the comparison. For the two example corpora studied in~\cite{Thelen2020} (a collection of German political documents and a collection of engineering research papers), the combination of Louvain modularity optimization with unadjusted edge weights results in the best topics when judged by human interpretability and coherence. More importantly, another insight emerging from the same study is that human interpretability can be significantly increased by presenting the topic terms in an appropriate order. The present paper is a direct continuation of~\cite{Thelen2020}, extending it in several directions: in addition to introducing new features for term extraction and topic presentation, we generalize the community detection procedure used, now including a resolution parameter that allows to tune the granularity of the topics, which is a novelty compared to all existing graph-based topic detection methods.

In passing we remark that network analysis is applied to document collections not only in the form of word co-occurrence networks, but also by studying co-author and citation networks, and both, too, have been exploited for topic discovery~\cite{Zeng2010, Guo2009}.

Term ranking will play an important role in our approach. On document level, term ranking is closely related to the problem of unsupervised key word extraction. This field is summarized in~\cite{Siddiqi2015, Firoozeh2020}. On corpus level, we are not aware of any method that is comparable to ours. However, there is a vague resemblance to the method for detecting {hot words} in microblogs described in~\cite{Yu2016}.

Word embeddings like Word2Vec~\cite{Mikolov2013} are a very efficient way of capturing the contextual information contained in large text collections for use in semantic text analysis. While we use pre-trained fastText embeddings~\cite{Bojanowski2017} for structuring and assessing the topics that we find by detection of term communities, other authors have used similar word embeddings directly for identifying topics through clustering in the embedding space~\cite{Butnaru2017} or indirectly for improving probabilistic topic models~\cite{Das2015}.

Vector space embedding strategies have recently been applied to the problem of community detection itself~\cite{Cavallari2019}. This is part of the very active research area of graph embeddings with its important applications in the context of knowledge graphs~\cite{Ji2020}. Considering the additional boost brought to graph-based algorithms by GPU hardware support~\cite{Tran2018}, it is timely to pay more attention to the potential of topic discovery via community detection.

\section{A Term-Community-Based Topic Discovery Method}
\label{sec:method}

It is a very intuitive idea that topics within a text corpus show up as patterns in the overall word usage in the corpus documents. Graphs as the mathematical structure for representing networks of entities (so called {nodes} or {vertices}) which are linked (through so called {edges}) are an obvious choice for formalizing this idea. What is less obvious is which of the many possible ways of transforming the corpus into a graph is the most effective one for the present purpose and how exactly topic-forming patterns can be identified.

In this section, we describe our particular choice which we found to be successful in the analyses of many text corpora.

\begin{figure}[H]
	\includegraphics[width=380bp]{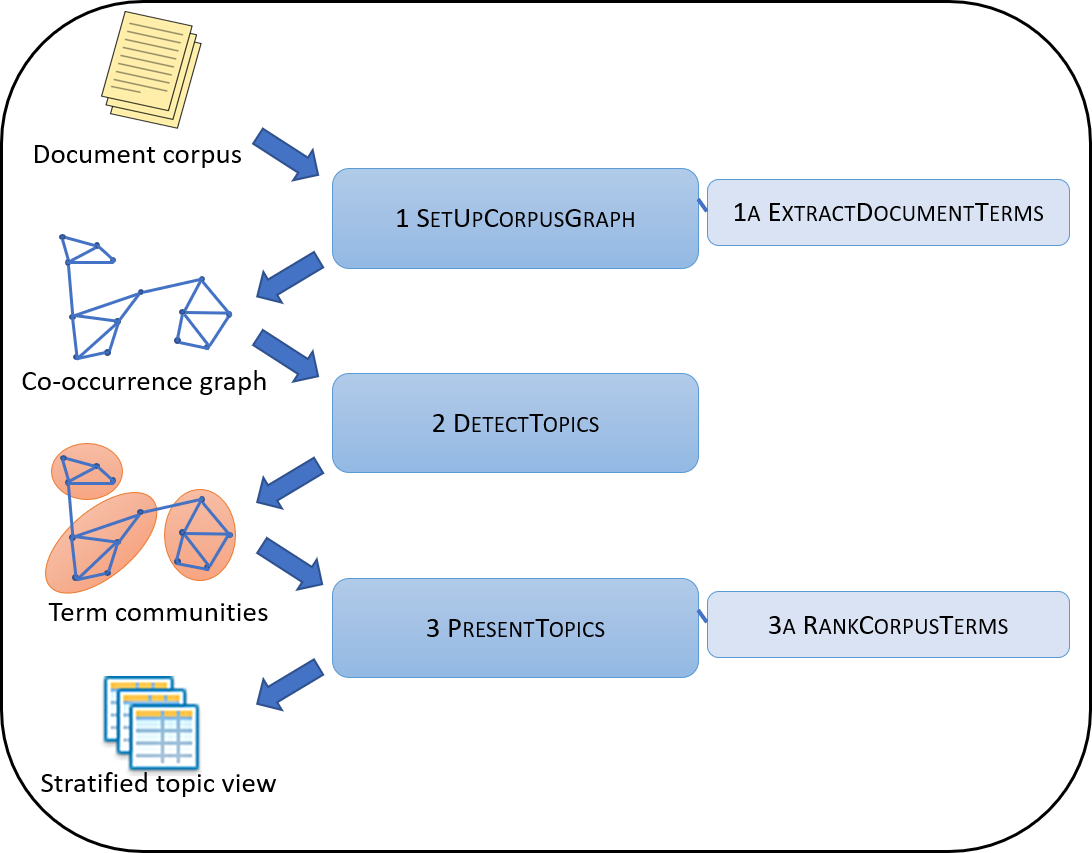}
	\caption{Workflow leading from a corpus of documents to a tabular presentation of topics. The various procedures involved in the workflow are described in detail in {Sections} %MDPI: please consider this suggested change. Similar highlights are same meaning . %AH: Change from Subsection to Section is okay here and in the other places.
 \ref{subsec:network}--\ref{subsec:presentation}.}
	\label{fig:workflow}
\end{figure}

Figure \ref{fig:workflow} shows how the method proceeds in three stages: first, a {corpus of text documents} is transformed into a {term co-occurrence network}. Second, topics are detected as {term communities} in this network. Third, the detected topics, which are initially unordered sets of terms, are presented with some two-dimensional structure, which we call \mbox{{stratified topic view}.}

In the following three subsections we will explain these stages in detail. For better overview, each subsection includes schematic pseudo code summarizations of the various procedures involved in the method (which are represented by blue boxes in \mbox{Figure \ref{fig:workflow}}). More detailed explanations, illustrative examples, and motivating comments are also presented.

One important matter is worth mentioning already now: within the general workflow of our method there is much room for variation in detail. A concrete implementation of the method requires choosing values for a dozen of parameters. Optimizing model parameters is particularly problematic in the present situation not only because of the extension of the parameter space but also because of the lack of a convincing target function that describes the model success, which ideally would measure the interpretability of the topics. We will come back to this point in Section \ref{sec:interpretability_comparison}.

We will give reasons for some of our choices which we found to be successful in the analyses of many text corpora. However, for several parameters we heuristically tried only a few values and chose the one which worked best on some manually assessed samples; we do not claim that we have tested all or even a big fraction of the conceivable alternatives. In principle, the parameters might be used in the future to further tune the results for \mbox{better performance.}

As mentioned before, we use a corpus of 2225 news articles from BBC news dating back to 2004 and 2005~\cite{Greene2006} as an example corpus for explaining and evaluating our method. The documents of this corpus each consist of a few hundred words.

\subsection{Setting up the Corpus Network}
\label{subsec:network}

Based on a corpus $\mathcal{D} = \{d_1, ..., d_N\}$ of $N$ text documents $d_i$, we define the corpus graph as a weighted undirected graph $G_p(\mathcal{D}) = (V_p(\mathcal{D}), E_p(\mathcal{D}), W_p(\mathcal{D}))$, consisting of a vertex set, an edge set, and an edge weight function. In this subsection we describe the various steps of the graph construction which are schematically depicted in Procedure 1 and Procedure \textsc{1a}. 

\begin{procedure}[h]
  \SetAlCapNameSty{textbfsc}
  \caption{1 SetUpCorpusGraph()}
  \DontPrintSemicolon
  \SetAlgoLined
  \KwIn{Corpus $\mathcal{D} = \{d_1, ..., d_N\}$ of $N$ text documents $d_i$ }
  \KwParameters{Reduction percentage $p$ with $0 < p \leq 100$}
  \KwOut{Term co-occurrence graph $G_p(\mathcal{D}) = (V_p(\mathcal{D}), E_p(\mathcal{D}), W_p(\mathcal{D}))$}
  $V_p(\mathcal{D})\leftarrow \{\}$\;
  $E_p(\mathcal{D})\leftarrow \{\}$\;
  \For{$n\leftarrow 1$ \KwTo $N$}{
    $((t_i, r_i), i=0,\dots,L)\leftarrow \textsc{ExtractDocumentTerms}(d_n, \mathcal{D})$ \emph{[Procedure \textsc{1a}]}\;
     \For {$j\leftarrow 1$ \KwTo $\left\lfloor p*L/100 \right\rfloor$}{
       $V_p(\mathcal{D})\leftarrow V_p(\mathcal{D}) \cup \{t_j\}$\;
       \For {$k\leftarrow 1$ \KwTo $j-1$}{
         \uIf {$e_{t_jt_k}\notin E_p(\mathcal{D})$}{
           $E_p(\mathcal{D})\leftarrow E_p(\mathcal{D}) \cup \{e_{t_jt_k}\}$\;
           $ W_p(\mathcal{D})(e_{t_jt_k}) \leftarrow 1$
         }
         \Else {
           $ W_p(\mathcal{D})(e_{t_jt_k}) \leftarrow  W_p(\mathcal{D})(e_{t_jt_k})+1$
         }
       }
     }
  }
\end{procedure}

\begin{procedure}[h]
  \SetAlCapNameSty{textbfsc}
  \caption{1a ExtractDocumentTerms()}
  \DontPrintSemicolon
  \SetAlgoLined
  \KwIn{Text document $d$ as part of a corpus $\mathcal{D}$ of documents}
  \KwOut{Sequence of document terms with rank values, $\mathcal{T}(d) := ((t_i, r_d(t_i))| i=0,\dots,l_d)$,\newline
                sorted by decreasing rank value: $r_d(t_i) \geq r_d(t_j)$ if $i < j$}          
  \emph{Recognize named entities in $d$.}\;
  \emph{Tokenize $d$ while treating compound named entities as single tokens.}\;
  \emph{Lemmatize tokens}\;
  \emph{Remove stop words.}\;
  \emph{Remove 1- and 2-character tokens, digits, exotic characters.}\;
  \emph{Keep only nouns, proper nouns, adjectives.}\;
  $l_d\leftarrow$ \emph{number of unique tokens left}\;
  $v(d) :=\{t_i|  i=0,\dots,l_d\}\leftarrow$ \emph{set of unique tokens left}\;
  \emph{Derive Markov chain on $v(d)$ with transition probabilities $P_{s, t}$ between terms $s, t \in v(d)$ as described in {Section} \ref{subsec:network} based on term positions and term neighborhood within $d$ as well as term statistics in $\mathcal{D}$.}\; 
  \emph{Compute the Markov chain's stationary distribution $r_d(t)$ for all $t\in v(d)$.}\;
\end{procedure}

The vertex set $V_p(\mathcal{D})$ is a certain subset of unique and normalized words or word combinations appearing in the corpus (which hereafter we will subsume under the name {terms}), where $p$ with $0 < p \leq 100$ is a parameter that controls the number of vertices. It is constructed by looping over all documents of the corpus (lines 4 to 6 in Procedure 1).

More specifically, for $p=100$ the terms that form the vertex set $V_{100}(\mathcal{D})$ result from a fairly standard text preparation pipeline (lines 2 to 8 in Procedure \textsc{1a}), consisting of tokenization, lemmatization (in order to consolidate several inflected forms of the same word), and the removal of stop words (frequent words with little meaning), short tokens (less than three characters), exotic characters, and tokens consisting mainly of digits. We also retain only nouns, adjectives, and proper nouns as usual in NLP tasks that focus on factual content. 

Applying the procedures listed in the previous paragraph to the original documents leads to single-word term vertices only. Yet, retaining compound terms that consist of several words as units in the corpus graph is a desirable enhancement of the method because it prevents the loss of meaning that is a consequence of splitting up a compound term into its components. Technically, we can include compound terms without changing the pipeline described above using a formal trick: before we put the documents into that pipeline, we connect the individual words of compound terms by underscores (e.g., ``department of homeland security'' becomes ``department\_of\_homeland\_security''). This renders the compound term a single token which survives the pipeline as a unit. However, identifying compound terms in documents is not an easy task. We experimented with various statistical and linguistic approaches. While it is possible to identify many meaningful combinations in that way, one also produces several nonsensical combinations which can create serious confusion in the results. Thus, we decided not to search for general compound terms but only for those which show up in named entity recognition (line 1 in Procedure \textsc{1a}). Concretely, we incorporate entities of the types {events}, {facilities}, {geographical and political entities}, {languages}, {laws}, {locations}, {nationalities or religious or political groups}, {organizations}, {persons}, {products}, and {works of art} which consist of 2, 3, or 4 words.

For all these document preparation steps, we use the Python library spaCy~\cite{Honnibal2020}. Working with the small language model en\_core\_web\_sm turned out to be sufficient for our purposes; using larger language models did not lead to significant changes in \mbox{the results.}

After these preparations every document $d_i$ has been stripped down to a collection of terms which still carry the main subject content of the document. $V_{100}(\mathcal{D})$ is the set of all unique terms remaining in the corpus. However, usually quite a few of these terms are of general nature and not important for the main message of the document. Having those terms in the corpus graph blurs its ability to represent thematic links. Working with a smaller subset of  $V_{100}(\mathcal{D})$, which we denote as $V_p(\mathcal{D})$ where $p<100$ is the percentage of terms retained, can prevent that effect as we will discuss in more detail in Section \ref{sec:influence}.

In order to judge about which terms to drop from the corpus graphs, a function for {document term ranking} is needed which produces a rank order of terms depending on their significance for the message of the document (lines 9 and 10 in Procedure \textsc{1a}). This is a well-known task in the context of key term extraction. The naïve solution would be to rank a term $t$  in document $d$ by its frequency $\Tf_d(t)$ in the document but it is well-known that this unjustly favors terms that tend to occur frequently, independent of the specific content of the document. The long-standing solution to this problem is to counterbalance the effect of overall frequent words by the {inverse document frequency},
\[
 \Idf(t) = \log \frac{N}{1 + \Df(t)}
\]
where $\Df(t)$ is the number of corpus documents that contain the term $t$, and to use \mbox{$\Tf_d(t) * \Idf(t)$} for term ranking in the document $d$.

However, this typical bag-of-word approach\textemdash where all words of a document are treated equally independent of their position\textemdash neglects the observation that important words of a document are usually not distributed evenly over the whole document. Rather, they tend to appear in groups, and for many document types it is also common that authors place especially many words that characterize the document content at the top of the document, notably in title, subtitle, and abstract. A term ranking method that takes these two observations into consideration is PositionRank~\cite{Florescu2017}, which is a modification of the TextRank method introduced in~\cite{Mihalcea2004} in analogy to the PageRank~\cite{BrinPage98} method for ranking within a network of linked web pages. 

In order to combine the advantages of the frequency arguments and the positional arguments for term ranking, we devised our own ranking method, posIdfRank~\cite{Hamm2019}, which works as follows: for a corpus document $d$, define a graph $g(d) = (v(d), e(d))$ (not to be confused with the corpus-wide graph $G_p(\mathcal{D})$) which has the set of unique terms of $d$ as vertex set $v(d)$, and an edge $\epsilon_{st} \in e(d)$  between two vertices $s$ and $t$ iff the terms $s$ and $t$ co-occur in a common window of  size $w$ (which is an adjustable parameter of the method), i.e., if there are at most $w-1$ words between $s$ and $t$.

We now consider a random walk on $g(d)$ in which the transition probability between two terms $s$ and $t$ is given by
\[
  P_{st} = \alpha \frac{\Idf(t) f_{st}}{\sum_{u\in v(d)} \Idf(u) f_{su}} + (1-\alpha)\frac{(1+\pos(t))^\beta \Idf(t)}{\sum_{u\in v(d)} (1+\pos(u))^\beta \Idf(u)}
\]
where $f_{st}$ counts how often in $d$ the terms $s$ and $t$ co-occur in a window of size $w$, and $\pos(t)$ counts on which position within the document the term $t$ appears first; $\alpha$ ($0\leq\alpha\leq1$) and $\beta$ ($\beta <0$) are two more parameters of the method. This process mimics a reader randomly scanning the document for important terms: from term $s$ he moves with probability $\alpha$ to another term $t$ in the vicinity (neighborhood of size $w$)\textemdash more likely reaching at terms which commonly stand close to $s$ and which do not appear in many documents of the corpus. However, with probability (1-$\alpha$) he jumps to some word which can be far away from $s$\textemdash then more likely to terms at the beginning of the document (where the preference of terms at the beginning is stronger the smaller one chooses $\beta$) and again to terms which do not appear in many documents of the corpus.

The long-term behavior of this random walk is characterized by its stationary distribution, a probability distribution $r_d(t)$ on $v(d)$. We regard this function as a useful document term ranking function: it has high values at terms that are frequently visited during the random walk. 

For the calculations in this paper, we fix the three parameters of the method at $\alpha=0.85$, $\beta=-0.9$ and $w=5$. These values were derived from a best fit with the manually assigned key words of the dataset of journal abstracts from~\cite{Hulth2003}. This also showed that the ranking is only weakly sensitive to moderate changes of the parameter values so that it is not critical that other training datasets result in slightly different optimal parameter values.  

We illustrate the considerations on document term ranking with the following example document from the BBC corpus:

\clearpage

\begin{mdframed}
{India widens access to telecoms} %MDPI: please check if it is box format. %AH: Format appears to be okay.

India has raised the limit for foreign direct investment in telecoms companies from 49\% to 74\%.

Communications Minister Dayanidhi Maran said that there is a need to fund the fast-growing mobile market. The government hopes to increase the number of mobile users from 95 million to between 200 and 250 million by 2007. ``We need at least \$20bn (£10.6bn) in investment and part of this has to come as foreign direct investment,'' said Mr Maran. The decision to raise the limit for foreign investors faced considerable opposition from the communist parties, which give crucial support to the coalition headed by Prime Minister Manmohan Singh. Potential foreign investors will however need government approval before they increase their stake beyond 49\%, Mr Maran said. Key positions, such as those of chief executive, chief technology officer and chief financial officer are to be held by Indians, he added.

Analysts and investors have welcomed the government decision. ``It is a positive development for carriers and the investment community, looking to take a longer-term view of the huge growth in the Indian telecoms market,'' said Gartner's principal analyst Kobita Desai. ``The FDI relaxation coupled with rapid local market growth could really ignite interest in the Indian telecommunication industry,'' added Ernst and Young's Sanjay Mehta. Investment bank Morgan Stanley has forecast that India's mobile market is likely to grow by about 40\% a year until 2007. The Indian mobile market is currently dominated by four companies, Bharti Televentures which has allied itself with Singapore Telecom, Essar which is linked with Hong Kong-based Hutchison Whampoa, the Sterling group and the Tata group.
\end{mdframed}

If one would simply go by the frequency of terms in the document, the 10 highest ranked terms would be:

\begin{mdframed}
investment, market, foreign, mobile, India, telecom, government, investor, chief, indian
\end{mdframed}

This list obviously does contain useful key terms of the document, but also very unspecific terms like ``market'' and ``chief''. In contrast, the top ranking according to $\Tf_d(t) * \Idf(t)$ would be:

\begin{mdframed}
Maran, investment, telecom, mobile, indian, foreign, India, investor, market, direct
\end{mdframed}

Here, the unspecific terms disappear or get shifted to lower positions. The very specific person name Maran, in contrast, appears at the top of the list.

Finally, the ranking according to posIdfRank results in the following top terms: 

\begin{mdframed}
India, telecom, investment, foreign, limit, Maran, direct, investor, mobile, Sanjay\_Mehta
\end{mdframed}

This list now favors important terms of the title and subtitle, and also brings up person names which stand close to other key terms in the document. Among the three ordered term lists, this is the one which condenses the content of the document best.

Now that we have a document term ranking function we can define the corpus vertex set $V_p(\mathcal{D})$: it is the result of keeping only the top $p$ percent (in the sense of the document term ranking function posIdfRank) of unique terms of each document (lines 5 and 6 of Procedure 1). 

In order to demonstrate the effect of this document reduction, we show what remains from the above example document if we keep only terms from $V_{50}(\mathcal{D})$:

\begin{mdframed}
India access telecom India limit foreign direct investment telecom Communications Minister mobile market government mobile investment foreign direct investment Maran decision limit foreign investor considerable opposition communist crucial coalition Minister Manmohan\_Singh foreign investor government Maran investor government decision investment indian telecom market Gartner principal Kobita\_Desai relaxation market indian telecommunication Ernst Young Sanjay\_Mehta investment bank Morgan\_Stanley India mobile market indian mobile market bharti\_televenture Essar hutchison\_whampoa Sterling
\end{mdframed}

Having established the vertex set, we now have to specify the edge set $ E_p(\mathcal{D})$ and the edge weight function $ W_p(\mathcal{D})$ (lines 7 to 14 in Procedure 1). The edges are supposed to connect terms that are likely to help in identifying thematic relationships, and the edge weight function should measure how significant this connection is. Among all conceivable options, the simplest turns out to be very effective already: two terms are connected if they appear together in at least one document and the weight of that connection is the number of documents in which both terms co-occur:
\[
 e_{st} \in E_p(\mathcal{D}) \iff  ( s, t \in  V_p(\mathcal{D}) \land \exists i\in \{1, \dots, N\}:  s, t \in d_i )  
\]
\[
  W_p(\mathcal{D})(e_{st}) = \# \{ i\in \{1, \dots, N\}| s, t \in d_i \}
\]

{S}%MDPI: Please check if indenting is required here. Similar highlights are same meaning. %AH: Indentation necessary because of start of a new paragraph. Similar in the other two places.
everal other authors who use network-based topic detection work with more restrictive and complicated edge definitions and weights, involving thresholds and $\Idf$-values in order to avoid noise produced by accidental or meaningless co-occurrences. We did not find this beneficial in our case as we avoided such type of noise already by reducing the vertex set $V_p(\mathcal{D})$ with percentages $p$ well below 100. 

We will go into more detail concerning the influence of the value of $p$ in Section \ref{sec:influence}. In the present section we work with $p=50$.

Following Procedures 1 and \textsc{1a}, the BBC corpus of 2225 documents leads to a graph $G_{50}(\mathcal{D_{\rm BBC}})$ that has 23859 vertices and 1896807 edges. The edge weights vary between 1 and 86 (with highest weight on the link between the terms {election} and {Labour}, meaning that 86 documents of the corpus contained both these terms).

\subsection{Detecting Topics as Communities}
\label{subsec:communities}

 In this subsection we describe how we detect topics in the term co-occurrence graph. The scheme for Procedure 2 outlines this stage of the workflow.

We first look into the details of its fundamental building block: community detection (line 2 of Procedure 2).
The aim of setting up the graph $G_p(\mathcal{D})$ was that in it topics of $\mathcal{D}$ would show up as communities of terms, where we refer to the network-theoretical concept of communities, which are\textemdash loosely speaking\textemdash groups of nodes that are densely connected within each group and sparsely connected with other groups. While this vague idea is quite intuitive, there are various non-equivalent options how to turn it into a quantifiable criterion for identifying communities, some of which we have mentioned \mbox{in Section \ref{sec:related_work}}. It is a priori not clear which criterion fits best for the language related purpose of establishing dense thematic connections. Comparative work in~\cite{Thelen2020} indicates that modularity maximization~\cite{Newman2006} is very well suited for achieving the goal of topic detection.

\newpage
%\vspace{6pt}
\begin{procedure}[h]
  \SetAlCapNameSty{textbfsc}
  \caption{2 DetectTopics()}
  \DontPrintSemicolon
  \SetAlgoLined
  \KwIn{Term co-occurrence graph $G_p(\mathcal{D}) = (V_p(\mathcal{D}), E_p(\mathcal{D}), W_p(\mathcal{D}))$ }
  \KwParameters{Resolution parameter $\gamma$\newline
                            Number of repetitions $N_{rep}$\newline
                            Number of concurrences $N_{con}\leq N_{rep}$\newline
                            Minimum size $S_{min}$}
  \KwOut{Set of topics $\{\hat{C_i}| i=1,\dots,\hat{k}\}$ where each topic is a set of terms $\hat{C_i}=\{t^{(i)}_j| j=1\dots m^{(i)}\}$}
  \For{$i\leftarrow 1$ \KwTo $N_{rep}$}{
    Mapping $c^{[i]}: V_p(\mathcal{D})\rightarrow \{1,\dots,k^{[i]}\}$ \emph{which assigns to each term a community number
    as obtained from \textsc{LeidenCommunityDetection}($G_p(\mathcal{D})$, $\gamma$) (greedy maximization of generalized modularity $\mathcal{Q}_\gamma$, see~\cite{Traag2019} and {Section} %AH: Changed in order to be consistent
    \ref{subsec:communities})}
  }
  $V\leftarrow V_p(\mathcal{D})$\;
  $i\leftarrow 1$\;
  \While{$V$ not empty}{
    \emph{$t\leftarrow$ element of $V$}\;
    \emph{remove $t$ from $V$}\;
    $S\leftarrow \{t\}$\;
    \ForEach{$s \in V$}{
      \If{$\# (\{i| c^{[i]}(s) = c^{[i]}(t) \ {\rm for}\ i=1,\dots,N_{rep}\})\geq N_{con}$}{
        $S\leftarrow S\cup\{s\}$\;
        \emph{remove $s$ from $V$}\;
      }
    }
    \If{$\# (S)\geq S_{min}$}{
      $\hat{C_i} \leftarrow S$\;
      $i\leftarrow i+1$\;
    }
  }
\end{procedure}

Given an undirected weighted graph $G=(V, E, W)$, modularity is a function which maps a partition $\mathcal{C} = \{C_j| j= 1, \dots, k \}$ of $V$ into vertex groups $C_j \subset V$ (with $\dot\cup_{j=1}^{k} C_j = V$) to a real number which measures how well the groups $C_j$ can be considered to be communities in the sense of high intra-group connectedness. It can be thought of as consisting of two parts. Denoting for a vertex $s\in V$ the group of $\mathcal{C}$ to which $s$ belongs by $c(s)$ and defining the total edge weight  $M = \frac{1}{2}\sum_{s, t \in V}W(e_{st})$ the first part,
\[
 \mathcal{I}(\mathcal{C}) = \frac{1}{2M} \sum_{s,t \in V}W(e_{st}) \delta_{c(s)c(t)} ,
\]
represents the fraction of all intra-group edge weights compared to the total edge weight of $G$, where we have used Kronecker's $\delta$ notation ($\delta_{ij}:=1$ for $i=j$ and $\delta_{ij}:=0$ for $i\neq j$) in order to express that $s$ and $t$ should be in the same group. The second part is the same fraction, but not for the graph at hand, $G$, but what one would expect for a random graph that has the same degree distribution (i.e., edge weight sum connected to each vertex) as $G$:
\[
 \mathcal{J}(\mathcal{C}) = \frac{1}{(2M)^2} \sum_{s,t \in V}K_s K_t \delta_{c(s)c(t)} ,
\]
with $K_s = \sum_{t\in V}W(e_{st})$, the edge weight sum (degree) at vertex $s$.

Modularity now is the difference between these two terms:
\[
 \mathcal{Q}_1(\mathcal{C}) =  \mathcal{I}(\mathcal{C}) -  \mathcal{J}(\mathcal{C}) .
\]

Maximizing the modularity therefore means finding a partition of the vertex set such that within the groups of this partition the fraction of intra-group edges is as big as possible when compared to the fraction of intra-group edges which one would expect in a similar random graph. At least intuitively, this translates well into what we are looking for: communities of terms that appear more often together in common documents than one would expect if the terms were randomly distributed. 

The above definition of modularity can be generalized to include a parameter $\gamma$ like follows~\cite{Reichardt2006}:
\[
 \mathcal{Q}_{\gamma}(\mathcal{C}) =  \mathcal{I}(\mathcal{C}) -  \gamma \mathcal{J}(\mathcal{C}) .
\]

{M}aximizing $\mathcal{Q}_{\gamma}(\mathcal{C})$ leads to coarser or finer communities depending on the value of $\gamma$: in the extreme situation that $\gamma=0$, the objective function is just $\mathcal{I}(\mathcal{C})$, and its maximum is obviously reached at the trivial solution $\mathcal{C} = \{V\}$, i.e., when the whole graph is considered to be one big community. In the other extreme, $\gamma \to \infty$, the objective is to minimize $\mathcal{J}(\mathcal{C})$, and this obviously happens when $\mathcal{C} = \{\{t\}| t \in V\}$, i.e., when each single vertex forms its own community such that the number of communities is equal to the number of vertices.

Varying $\gamma$ in the generalized modularity $\mathcal{Q}_{\gamma}(\mathcal{C})$ makes it possible to find communities of variable granularity, and therefore $\gamma$ is called the resolution parameter.

Computationally, maximizing the (generalized) modularity is known to be a non-deterministic polynomial-time hard problem, which means that there are no efficient algorithms that guarantee an optimal solution. However, several efficient heuristic algorithms are known for producing potentially suboptimal but useful solutions.

Here, we use the recently published Leiden algorithm~\cite{Traag2019}, which is an improvement of the very popular Louvain algorithm~\cite{Blondel2008} for maximizing modularity. Starting from the extreme partition $\{\{t\}| t \in V\}$ where each vertex forms its own community, the Leiden algorithm first visits the nodes in a random order and tries greedily to shift nodes to other communities in a way that offers the biggest modularity increases. In a second step (which is the main improvement compared to the Louvain algorithm), the community partition is refined in a way that produces well-connected communities. After that, an aggregated graph is formed which contains the refined communities of the original graph as vertices. The whole procedure is repeated for the aggregated graph, and this is iterated until no further modularity increase can be achieved.

For our calculations we use the implementation of the Leiden algorithm in the Python version of the library igraph~\cite{Csardi2006}.

Applying the Leiden algorithm with the standard resolution parameter $\gamma = 1$ to the example graph $G_{50}(\mathcal{D_{\rm BBC}})$ results in 8 to 14 term communities, depending on the run. Each run may produce a different number of communities because the algorithm follows a non-deterministic heuristic. Therefore, not only the number but also the terms of the communities, which are only approximations to the optimal community partitioning, may change in each run. However, closer inspection shows that the differences between the runs are not large. In particular, if a run produces more than 8 communities, the additional communities are strikingly smaller than the biggest 8 communities. This is an indication that in some runs the greedy algorithm fails to assign some terms to any of the dominant communities and leaves them in some small residual communities.

While these odd cases are reasonably easy to detect, it is nevertheless better to remove those undecisive terms completely from the picture. Therefore, for the actual topic detection we repeat the community detection step $N_{rep}$ times (lines 1 to 3 in Procedure 2) and consider only those sets of terms which end up together in the same Leiden community in at least $N_{con}$ of the runs (lines 10 to 15 in Procedure 2). Finally, of those sets we retain only the ones that contain a minimum number of $S_{min}$ terms (lines 16 to 19 in Procedure 2). We found that the remaining sets of terms form stable topics if we set $N_{rep}=20$, $N_{con}=15$ and $S_{min}$ to 10\% of the terms that one would expect in a topic if all topics were of equal size.

Following Procedure 2, 22426 terms among the 23859 vertices of $G_{50}(\mathcal{D_{\rm BBC}})$ (or 94\%) can be assigned to one of 8 stable topics. The biggest of these topics contains 4657 terms, the smallest 214 terms. 

Based on the assignment of terms to topics, we can also determine which document of the corpus is concerned with which topics. The connection is made by counting the relative number of topic terms within a document: if $f_i(d)$ is the number of terms in document $d$ that belong to the topic cluster $C_i$, then $\frac{f_i(d)}{\sum_{j=1}^{k}f_j(d)}$ is a good indicator of the importance of topic $i$ for that document. While one can argue that here a probabilistic topic model would offer a sounder method for calculating the topic share per document, we found that a simple count of topic terms works very well as we will show when we use the term community topics for document classification in Section \ref{sec:influence}.

Like with other topic detection approaches, the method results in a list of terms that characterize a topic, but the actual interpretation of what that topic is about, is left to human evaluation. It is certainly a difficult task to look at hundreds or thousands of terms, having to make sense of what topic might be encoded in them. Probabilistic topic models that produce topics as probability distributions on terms have an advantage here at first glance: the topic terms can be sorted by their probability, and one usually looks only at the 10 to 30 most probable terms.

In the next subsection we will explain how we suggest to look at big sets of topic terms in a way that facilitates interpretation in the absence of a probability distribution on the terms.

\subsection{Topic Presentation}
\label{subsec:presentation}

In this subsection we explain how we present the topic terms in an informative way that eases topic interpretation. The relevant steps are schematically outlined in Procedure 3 and Procedure \textsc{3a}. 

\begin{procedure}
  \SetAlCapNameSty{textbfsc}
  \caption{3 PresentTopic()}
  \DontPrintSemicolon
  \SetAlgoLined
  \KwIn{Topic $\hat{C}=\{t_j| j=1\dots m\}$ obtained from \textsc{DetectTopics} \emph{[Procedure \textsc{2}]}\newline
             Corpus term ranking function $r(t)$ for $t\in V_p(\mathcal{D})$ obtained from \textsc{RankCorpusTerms} \emph{[Procedure \textsc{3a}]}}
  \KwOut{Tabular presentation $(t_{\alpha\beta})$ of topic terms}
  \emph{Define distance between $s,t\in \hat{C}$ as Euclidean distance between \textsc{FastTextEmbedding}($s$) and \textsc{FastTextEmbedding}($t$), see~\cite{Bojanowski2017}}\;
  \emph{Based on that distance function, partition $\hat{C}$ into $K$ clusters:}\newline
  $\{\tilde{C}_{\kappa}| \kappa=1,\dots K\}\leftarrow$ \textsc{AgglomerativeClustering}($\hat{C}$) \emph{\cite{Scikit2011}}\;
  \For{$\kappa\leftarrow 1$ \KwTo $K$}{
    \emph{Sort the $B_{\kappa}$ terms in $\tilde{C}_{\kappa}$ in descending order with respect to the corpus term ranking function:}\newline
    $(\tilde{t}_{\kappa\beta} | \beta=1,\dots, B_{\kappa})$ such that $r(\tilde{t}_{\kappa\iota})\geq r(\tilde{t}_{\kappa\lambda}))$ if $\iota < \lambda$.\;
  }
  \emph{Sort the $K$ clusters $\tilde{C}_{\kappa}$ in decreasing order with respect to the term ranking function applied to its first terms $\tilde{t}_{\kappa 1}$.}\;
  \For{$\kappa\leftarrow 1$ \KwTo $K$}{
    \emph{$\alpha(\kappa) \leftarrow$ position of $\tilde{C}_{\kappa}$ in that sorted sequence}\;
    \For{$\beta\leftarrow 1$ \KwTo $B_{\kappa}$}{
      $t_{\alpha(\kappa)\beta} \leftarrow \tilde{t}_{\kappa\beta}$.\;
    }
  }
\end{procedure}

\begin{procedure}
  \SetAlCapNameSty{textbfsc}
  \caption{3a RankCorpusTerms()}
  \DontPrintSemicolon
  \SetAlgoLined
  \KwIn{Set of sequences of document terms with rank values, $\{\mathcal{T}(d) := ((t_i, r_d(t_i))| i=0,\dots,l_d) |d\in\mathcal{D}\}$, obtained from \textsc{ExtractDocumentTerms} 
             \emph{[Procedure \textsc{1a}]}}
  \KwOut{Corpus term ranking function $r(t)$ for $t\in V_p(\mathcal{D})$}
  \ForEach{$d \in \mathcal{D}$}{
    \For {$i\leftarrow 1$ \KwTo $l_d$}{
      \emph{Replace continuous rank value $r_d(t_i)$ of $i$-th document term by a discretized rank value $q_d(t_i)$ as described in Section \ref{subsec:presentation}.}
    }
  }
  \ForEach{$t \in V_p(\mathcal{D})$}{
    \emph{$r(t) \leftarrow \textsc{BayesianAverage}(\{ q_d(t) | d\in\mathcal{D}\})$ [described in Section \ref{subsec:presentation}]}\;
  }  
\end{procedure}

\newpage

Looking in a random order at the hundreds or thousand terms which constitute a topic is certainly not helpful for grasping its meaning. It would be best to look at the most characteristic terms first; what we need is a term ranking. In the context of network preparation we have already established a {document term ranking} $r_d(t)$. However, what is required now is a {corpus term ranking}. The former one can only decide which terms are the most characteristic ones for a certain document $d$, but now we need a more global ranking function $r(t)$, independent of $d$. For a term $t$, the document term ranking function $r_d(t)$ is only defined for those documents $d$ which contain $t$. Since the document frequency $\Df(t)$ varies a lot with $t$ it would not be fair to simply take the arithmetic average of the existing values of $r_d(t)$. This resembles star ratings in recommender systems where typically some items have been rated many times and some items have only one rating. There one uses a Bayesian average for ranking~\cite{Yang2013}.

In order to transfer that solution to the problem of corpus term ranking, we first introduce a discretized version $q_d(t)$ of the document term ranking $r_d(t)$ (lines 1 to 5 in Procedure \textsc{3a}). This is because we do not claim that the document term ranking is sufficiently exact to measure the importance of terms continuously but rather that it is a good base for grouping terms into sets of more or less important terms. Let $T(d) = (t_1, \dots, t_{m_d})$ be an ordered sequence of the terms in document $D$, $t_i \in d$ for $i=1,\dots,m_d$ and $r_d(t_i) > r_d(t_j)$ if $i<j$. Then we divide $T(d)$  into $A$ parts $T_a(d)$, $a=1,\dots,A$, of equal lengths $[m_d/A]$. We also introduce a cut-off value $K\leq A$; $A$ and $K$ are parameters which can be adjusted so as to result in a ranking that works well.

\[
 q_d(t) = b  \iff  t \in T_{K+1-b}(d) \quad{\rm for}\quad b=1,\dots,K
\]
and $q_d(t) = 0$ otherwise. After some experimentation we fixed $A=20$ and $K=3$; this means that the top 5\% of terms in a document get the discretized rating 3, the next 5\% the rating 2, and the third 5\% the rating 1.

Based on $q_d(t)$, we calculate the corpus term ranking function $r(t)$ as the following Bayesian average (lines 6 to 8 in Procedure \textsc{3a}):
\[
 r(t) = \frac{C*L + \sum_{d: t\in d}q_d(t)}{C + \Df(t)}
\]
with $C = \frac{\sum_{s\in V_{100}(\mathcal{D})} \Df(s)}{\#(V_{100}(\mathcal{D}))}$, which is the document frequency averaged over all terms, and $L = \frac{K(K+1)}{2A}$, which is the mean $q_d(t)$ ranking of all terms.

With the function $r(t)$ we can sort all terms of the corpus. 

The ten terms which ranked highest as the most specific terms for the corpus \mbox{$\mathcal{D_{\rm BBC}}$ are:}

\begin{mdframed}
Yukos, Holmes, UKIP, Fiat, Blunkett, howard, Yugansk, Kenteris, Parmalat, Wenger
\end{mdframed}

The ten terms ranked lowest\textemdash the least specific ones\textemdash are:

\begin{mdframed}
year, place, time, month, spokesman, recent, week, Tuesday, long, second
\end{mdframed}

Now we can present the terms that form a topic in the order of decreasing values of $r(t)$. The examples in Table \ref{tab:randomvsordered} show how advantageous this is. Both columns show terms from the same topic found by community detection in $G_{50}(\mathcal{D_{\rm BBC}})$ with resolution parameter $\gamma=1$. The topic comprises a total of 4080 terms. 

The left column of the table shows just 18 randomly picked topic terms. Guessing the topic’s theme from these mostly general terms, like ``contribution'' or ``flexible'', and less familiar person names or book titles is difficult. More generally, the chance that random samples include the most characteristic terms of the topic is low. In contrast, the right column shows the 18 topic terms with highest values of $r(t)$. Most of these terms are prominent politicians (e.g., ``Blair'', ``Blunkett''), political parties (e.g., ``UKIP'', ``lib\_dem'') and other well-known entities (e.g., ``Speaker'', ``migrant'') from---not only, but mainly, British---politics. This makes the topic's main theme immediately evident to \mbox{domain experts.}

\begin{specialtable}[H]
	\caption{Comparison of selected terms of the same topic. The left column picks some random terms; the right column shows the terms with highest r value.}\label{tab:randomvsordered}
\setlength{\cellWidtha}{\columnwidth/2-2\tabcolsep+0.0in}
\setlength{\cellWidthb}{\columnwidth/2-2\tabcolsep+0.0in}
\scalebox{1}[1]{\begin{tabularx}{\columnwidth}{
>{\PreserveBackslash\centering}m{\cellWidtha}
>{\PreserveBackslash\centering}m{\cellWidthb}}

\toprule
      \textbf{Random Order} & \textbf{r-Sorted Order}\\
\midrule
	contribution &       UKIP \\
	 supporter &   Blunkett \\
	Czechoslovakia &     howard \\
	unwinnable &       Silk \\
 	sentencer &     Kilroy \\
 	Skidelsky &    Speaker \\
	     tract & Guantanamo \\
	registration &    Kennedy \\
	Julian\_Lewis &      Hague \\
	  flexible &    hunting \\
	elizabeth\_blackman &      Blair \\
	Lawless\_World &      blair \\
	   feeling &     quango \\
	      Eady & Mr\_Blunkett \\
	   Moazzam & Guantanamo\_Bay \\
	 secretary &    lib\_dem \\
	     Brady &    migrant \\
	understanding &  Goldsmith \\
\bottomrule

\end{tabularx}} 
\end{specialtable}

However, looking at only one or two dozen of 4080 identified topic terms wastes a lot of information, and we suggest to look at many more terms when interpreting topics in order to achieve a proper assessment. In order to facilitate an overview {of} %AH: correction
many topic terms, we add,  next to the specificity ranking $r(t)$, another structuring criterion to the set of topic terms by grouping them into clusters of semantically related terms. Here, we make use of pretrained fastText embeddings~\cite{Bojanowski2017}. The fastText approach belongs to the semantic work embedding methods through which words can be mapped to a moderately low-dimensional Euclidean vector space in a way that semantic closeness translates into small metric distances. The pre-trained fastText models are shallow neural nets with output vectors of dimension 300 that were trained on huge text collections from Wikipedia and Common Crawl, using the so called CBOW task of predicting a word by its surrounding words. In distinction to its predecessor Word2Vec, fastText internally does not work on the level of words but on the level of its constituting character n-grams. In the present context this offers two advantages: first, this mapping works even on words which do not appear in the Wikipedia and Common Crawl collections; second, word variations due to spelling mistakes or imperfect lemmatization usually end up close to each other in the vector space representation. 

If we now take the vector representations of the topic terms, we can use any metric-based clustering method for finding groups of semantically related words, or, more precisely, of words whose components have been seen frequently together in huge text collections. After some experimentation, we decided to use hierarchical clustering in its scikit-learn~\cite{Scikit2011} implementation AgglomerativeClustering with distance threshold 1 \mbox{(lines 1 and 2} in Procedure 3).

We show these semantic groups of terms, which we call semantic strata, rather than single terms, when we present the topics. We order the terms within each stratum by their $r(t)$ value (lines 3 to 5 in Procedure 3) and the strata by the value of $\max r(t)$ for the terms $t$ per stratum (lines 6 to 12 in Procedure 3). As a result, we have a two-dimensional order in the topic terms: one dimension ranking the specificity and one dimension depicting semantic relations. 

For topic evaluation we produce large sheets with the topic terms structured in the stratified way described above. In Table \ref{tab:stratified}, we show only the beginning of such a sheet for better comprehension. The rows depict the strata in which the $r$-sorted top terms of the topic get  accompanied by semantically related terms, or, in the case of person names, by persons who usually appear in a common context.

% start a new page without indent 4.6cm
%\clearpage
\end{paracol}
\nointerlineskip
\begin{specialtable}[H] 
\widetable

	\caption{Principle of stratified topic view. Terms with high r values are complemented through other topic terms from their fastText embedding cluster.}\label{tab:stratified}
\setlength{\cellWidtha}{\columnwidth/4-2\tabcolsep+0.0in}
\setlength{\cellWidthb}{\columnwidth/4-2\tabcolsep+0.0in}
\setlength{\cellWidthc}{\columnwidth/4-2\tabcolsep+0.0in}
\setlength{\cellWidthd}{\columnwidth/4-2\tabcolsep+0.0in}
\scalebox{1}[1]{\begin{tabularx}{\columnwidth}{
>{\PreserveBackslash\centering}m{\cellWidtha}
>{\PreserveBackslash\centering}m{\cellWidthb}
>{\PreserveBackslash\centering}m{\cellWidthc}
>{\PreserveBackslash\centering}m{\cellWidthd}}

\toprule

	    {UKIP} %MDPI: Please add table header. Similar highlights are same meaning %AH: The columns of this table must NOT have headings as they are supposed to give an example of how the stratified topic view actually is presented. The individual columns do not have any specific meaning. Rather, it is a two-dimensional representation of a term community. Similarly in the other marked table.
  &            &            &            \\
\hline

	  Blunkett &  Mandelson &   Miliband &  Heseltine \\
	\hline
	    howard &            &            &            \\
	\hline
	      Silk &            &            &            \\
	\hline
	    Kilroy &            &            &            \\
	\hline
	   Speaker &            &            &            \\
	\hline
	Guantanamo &   detainee &   prisoner &  captivity \\
	\hline
	   Kennedy &   Moynihan &            &            \\
	\hline
	     Hague &            &            &            \\
	\hline
	   hunting &   pheasant &    Hunting &            \\
	\hline
	     Blair &     Blairs &            &            \\
	\hline
	     blair &            &            &            \\
	\hline
	    quango &    Quangos &            &            \\
	\hline
	Mr\_Blunkett & David\_Blunkett & David\_Miliband & david\_miliband \\
	\hline
	Guantanamo\_Bay &   mr\_hague & the\_british\_national\_party & the\_world\_economic\_forum \\
	\hline
	   lib\_dem & White\_Paper & Royal\_Mail & Upper\_House \\
	\hline
	   migrant &    refugee &            &            \\
\bottomrule

\end{tabularx}} 
\end{specialtable}
\begin{paracol}{2}
%\linenumbers
\switchcolumn

\vspace{-6pt}

\section{The Influence of the Resolution Parameter and of the Reduction Percentage}
\label{sec:influence}

In this and the following section, we present concrete observations derived from working with the BBC corpus. The general aim is to study the influence of modeling decisions and parameters on structure, interpretability, and applicability of the \mbox{detected topics. }

All topic term sheets produced in the way described in Section \ref{sec:method}, with various values of the parameters $\gamma$ and $p$ discussed in the following, were given to three evaluators from the social sciences with varying backgrounds in political science, economics, and sociology. Each topic $\hat{C}_i$ was assessed independently by two of them: the task was to interpret the topic and to find an appropriate label for it. The evaluators also graded the interpretability of the topic from score 1 (=hardly possible to make sense of the topic) to score 5 (=topic recognizable without any doubts). The third evaluator compared the two previous evaluations and prepared a meta evaluation. In the majority of cases, the two first evaluators gave identical or nearly identical labels, which were almost always approved by the meta evaluator. In most other cases, it was possible to agree on a consensus label; in very few cases, the evaluators found the topics unidentifiable. 

Denoting the interpretability score given by evaluator number $j (j\in\{1, 2, 3\})$ for topic $\hat{C}_i$ by $z_j(\hat{C}_i)$, the average of the three evaluator scores, $z(\hat{C}_i) = \frac{1}{3} \sum_{j=1}^3 z_j(\hat{C}_i)$, was taken as final score for this topic.

For measuring the interpretability of a whole set of $k$ topics, $\{\hat{C}_I| i=1,\dots k \}$, we use the following aggregated indicators: the mean score of all topics, $\bar{z} = \frac{1}{k} \sum_{i=1}^k z(\hat{C}_i)$ assesses the overall average interpretability of the topics. The number 
\[
  k^+=\#\{i| z(\hat{C}_i)=\{4,5\}\}
\]
counts how many topics were found easy to interpret. Likewise,
\[
  k^-=\#\{i| z(\hat{C}_i)=\{1,2\}\}
\]
is the number of topics that were assessed as problematic in interpretation. Obviously, it is not necessary to separately count the number of topics with medium score 3 as this is then already known to be $k-k^+-k^-$.

The numbers $k^+$ and $k^-$ supplement the average quality information about topic sets  inherent in the mean score $\bar{z}$ in that they allow to recognize situations where only a few topics with poor interpretability---potentially outliers---lower the mean score, or vice versa.

{Excel files containing term lists, expert given labels, and topic distributions are available online as Supplementary Material}.

\subsection{Varying the Resolution}

Next, we study the influence of the resolution parameter $\gamma$ at the example of the BBC corpus, rank-reduced to 50\% of its terms.

Table \ref{tab:varygamma} shows the resulting number $k$ of topics for different values of $\gamma$; some of which were chosen for later comparison in Section \ref{sec:interpretability_comparison}. As expected, $k$ rises with $\gamma$. The table also shows that the evaluators generally rated topic interpretability, expressed through the average score $\bar{z}$, as high. However, there is a clear trend that interpretability declines with higher resolutions. 

\begin{specialtable}[H]
	\caption{Evaluation of topics produced from $G_{50}(\mathcal{D_{\rm BBC}})$ with various values of the 
	resolution parameter $\gamma$. The table shows for each resolution the number $k$ of topics produced. 
	The numbers $k^+$ and $k^-$ indicate how many topics the evaluators found easy to interpret (score 5 or 4) 
	and hard to interpret (score 1 or 2), respectively; $\bar{z}$ is the mean score of all topics of that resolution.}	\label{tab:varygamma}

	\setlength{\cellWidtha}{\columnwidth/5-2\tabcolsep+0.0in}
\setlength{\cellWidthb}{\columnwidth/5-2\tabcolsep+0.0in}
\setlength{\cellWidthc}{\columnwidth/5-2\tabcolsep+0.0in}
\setlength{\cellWidthd}{\columnwidth/5-2\tabcolsep+0.0in}
\setlength{\cellWidthe}{\columnwidth/5-2\tabcolsep+0.0in}
\scalebox{1}[1]{\begin{tabularx}{\columnwidth}{
>{\PreserveBackslash\centering}m{\cellWidtha}
>{\PreserveBackslash\centering}m{\cellWidthb}
>{\PreserveBackslash\centering}m{\cellWidthc}
>{\PreserveBackslash\centering}m{\cellWidthd}
>{\PreserveBackslash\centering}m{\cellWidthe}}

\toprule

	  \boldmath{$\gamma$} &   \boldmath{$k$} & \boldmath{$k^+$} &  \boldmath{$k^-$} &   \boldmath{$\bar{z}$} \\
	\midrule
	 0.80 &   5 &   5 &   0 &  5.0 \\
	 1.00 &   8 &   8 &   0 &  4.8 \\
	 1.07 &  10 &  10 &   0 &  4.8 \\
	 1.37 &  19 &  18 &   0 &  4.7 \\
	 1.50 &  27 &  25 &   1 &  4.5 \\
	 2.00 &  58 &  44 &   6 &  4.2 \\
	 2.50 &  89 &  67 &  14 &  3.8 \\
\bottomrule

\end{tabularx}} 
\end{specialtable}

Up to resolution $\gamma=1.5$, there are hardly any topics that were difficult to interpret, but at $\gamma=2.5$, 15\% of the topics were found to be problematic. Nevertheless, it is remarkable that the method succeeds in producing $k^+=67$ clearly interpretable topics.

We will come back to issues of interpretability in Section \ref{sec:interpretability_comparison} but first discuss content aspects of increasing the resolution. We start with looking closer at the topics at resolution~\mbox{$\gamma=0.8$}.

The BBC corpus comes with a classification into 5 broad classes: Business, Entertainment, Politics, Sport, Tech. Resolution $\gamma=0.8$ produces 5 topics, labeled by the evaluators as follows: Sports, Music \& films, Technology, Politics (UK interests based), Economy. The congruence between classes and topics is obvious on the level of labels. In order to see how well this extends to the document level, we calculate the topic shares in each document. The topic with the highest share we call the dominant topic of the document. In this way we compile the crosstable Table \ref{tab:cross} between preassigned classes and detected \mbox{dominant topics.}

\begin{specialtable}[H]
	\caption{Crosstable between preassigned classes and detected dominant topics ($\gamma=0.8$) for all documents of the BBC corpus.}\label{tab:cross}
\setlength{\cellWidtha}{\columnwidth/6-2\tabcolsep+0.0in}
\setlength{\cellWidthb}{\columnwidth/6-2\tabcolsep-0.1in}
\setlength{\cellWidthc}{\columnwidth/6-2\tabcolsep+0.4in}
\setlength{\cellWidthd}{\columnwidth/6-2\tabcolsep-0.1in}
\setlength{\cellWidthe}{\columnwidth/6-2\tabcolsep-0.1in}
\setlength{\cellWidthf}{\columnwidth/6-2\tabcolsep-0.1in}
\scalebox{1}[1]{\begin{tabularx}{\columnwidth}{
>{\PreserveBackslash\centering}m{\cellWidtha}
>{\PreserveBackslash\centering}m{\cellWidthb}
>{\PreserveBackslash\centering}m{\cellWidthc}
>{\PreserveBackslash\centering}m{\cellWidthd}
>{\PreserveBackslash\centering}m{\cellWidthe}
>{\PreserveBackslash\centering}m{\cellWidthf}}

\toprule

	\textbf{Topic Class} &  \textbf{Economy} &  \textbf{Music} \& \textbf{Films} &  \textbf{Politics} &  \textbf{Sports} &  \textbf{Technology} \\
	\midrule
	Business      &      473 &            4 &        18 &       3 &          12 \\
	Entertainment &        7 &          346 &        12 &       3 &          18 \\
	Politics      &       18 &            1 &       396 &       0 &           2 \\
	Sport         &        1 &            1 &         1 &     507 &           1 \\
	Tech          &        5 &            4 &        11 &       3 &         378 \\
\bottomrule

\end{tabularx}} 
\end{specialtable}

The corresponding classification statistics is shown in Table \ref{tab:cstat}. However, what we have in the detected topic distribution is more than a simple classifier, as we do not only learn which is the dominant topic of a document but also what other topics are visible in a document. For instance, the example document about foreign investment in Indian telecom presented in Section \ref{sec:method} belongs to the class Business. However, in the topic term assignment, while Economy is the dominant topic with a share of 59\%, there is also a share of 25\% Politics and of 14\% Technology, which is a reasonable topic composition for the document. In fact, several of the few cases which were misclassified based on their dominant topic were borderline articles with two similarly strong appropriate topics. The following titles give some examples: the article ``News Corp eyes video games market'' belongs to class Business, but Technology was detected as dominant topic. The Entertainment article ``Ethnic producers face barriers'' has Politics as dominant topic. The article ``Arsenal may seek full share listing'' is in class Business, but the topic Sports dominated here.

\begin{specialtable}[H]
	\caption{Classification statistics for predicting preassigned classes by detected dominant topics ($\gamma=0.8$). Precision is the fraction of true positives among all positive
	predictions, recall is the fraction of true positives compared to all actual class members, and the f1-score is their harmonic mean. Values are given for each class separately
	and as an average weighted according to the sizes of the classes.}\label{tab:cstat}
\setlength{\cellWidtha}{\columnwidth/4-2\tabcolsep+0.0in}
\setlength{\cellWidthb}{\columnwidth/4-2\tabcolsep+0.0in}
\setlength{\cellWidthc}{\columnwidth/4-2\tabcolsep+0.0in}
\setlength{\cellWidthd}{\columnwidth/4-2\tabcolsep+0.0in}
\scalebox{1}[1]{\begin{tabularx}{\columnwidth}{
>{\PreserveBackslash\centering}m{\cellWidtha}
>{\PreserveBackslash\centering}m{\cellWidthb}
>{\PreserveBackslash\centering}m{\cellWidthc}
>{\PreserveBackslash\centering}m{\cellWidthd}}

\toprule

	{} &  \textbf{Precision} &    \textbf{Recall} &  \textbf{f1-Score} \\
	\midrule
	Business      &   0.938 &  0.927 &  0.933 \\
	Entertainment &   0.972 &  0.896 &  0.933 \\
	Politics      &   0.904 &  0.950 &  0.926 \\
	Sport         &   0.983 &  0.992 &  0.987 \\
	Tech          &   0.920 &  0.943 &  0.931 \\
	\midrule
	weighted avg  &   0.945 &  0.944 &  0.944 \\
	\bottomrule

\end{tabularx}} 
\end{specialtable}
%\vspace{-6pt}

We are now interested in what topics show up when we increase the resolution parameter $\gamma$. Resolution $\gamma=1$ results in 8 topics. The heat map in Figure \ref{fig:hm080to100} shows how the topic terms of these 8 topics\textemdash corresponding to the 8 rows\textemdash are distributed within the 5 topics of lower resolution $\gamma=0.8$\textemdash corresponding to the 5 columns. While the 5 original topics basically persist, 3 of them have smaller spin-offs: the general Sports topic gave rise to a new Athletics topic, from the Music \& films topic a new TV topic splits off, and the Technology topic forks into a new (Video) Gaming topic.

\begin{figure}[H]
%	\centering
	\includegraphics[width=380bp]{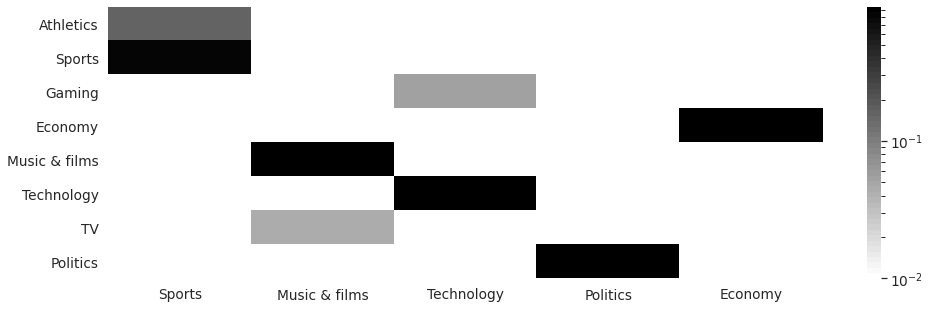}
	\caption{Heat map showing how topic terms of the 5 $\gamma=0.8$ topics (columns) get distributed on 8 topics at $\gamma=1.0$ (rows).}
	\label{fig:hm080to100}
\end{figure}

This phenomenon of small topics splitting off from a big one is a typical pattern contributing to the increasing number of topics at higher values of $\gamma$. The comparison of resolution $\gamma=1.37$ with 19 topics and $\gamma=1.0$ with 8 topics in Figure \ref{fig:hm100to137} shows further examples. The topic Politics decomposes into the big topic UK politics and small topics Terrorism, Euroscepticism, and Nutrition \& health. However, two more topics have significant contributions from the former Politics topic: Labour, which has also input from the former Economy topic, and Cybersecurity, which is primarily fed by the former \mbox{topic Technology.}

\begin{figure}[H]
	%\centering
	\includegraphics[width=380bp]{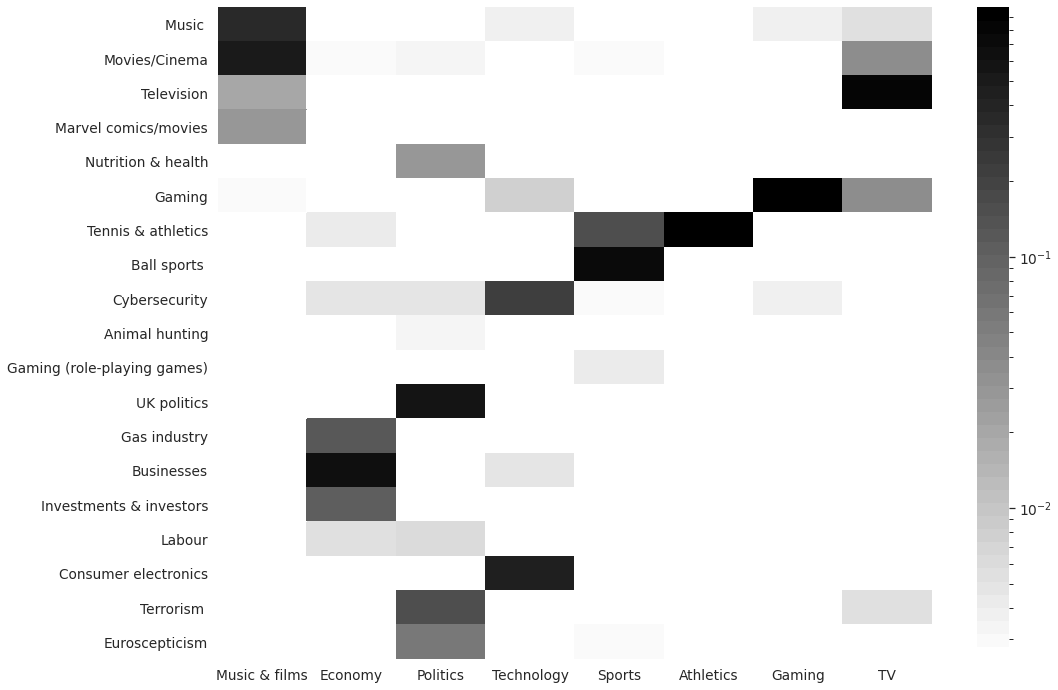}
	\caption{Heat map showing how topic terms of the 8 $\gamma=1.0$ topics (columns) get distributed on 19 topics at $\gamma=1.37$ (rows).}
	\label{fig:hm100to137}
\end{figure}

Another example illustrates that direct topic division is not the only way in which topics of higher resolution emanate. The high-resolution topics Music, Movies/cinema, Television, Marvel comics/movies altogether arise from a joint restructuring of the two coarser topics Music \& films and TV (and some faint influences of other topics).

Yet another form of recombination can be seen in the class Sport. At $\gamma=1$, it comprises two topics: Sports and Athletics. At $\gamma=1.37$, there are still only two sport topics: ball sports (with an emphasis on football and rugby) and a combined topic \mbox{Tennis \& athletics}. This means that here the higher resolution splits off Tennis from the general Sports topic but immediately combines it with the Athletics topic.

Going from $\gamma=1.37$ to $\gamma=1.5$, we end up with 27 topics (see Figure \ref{fig:hm137to150}). Here, remarkable developments are that the business and investment topics give now rise to 4 related topics, 2 of them now getting into regional or sector details (Development Asia and Aviation). Furthermore, the sport topics get more specific: Football and Rugby are separate \mbox{topics now.}

\begin{figure}[H]
%	\centering
	\includegraphics[width=380bp]{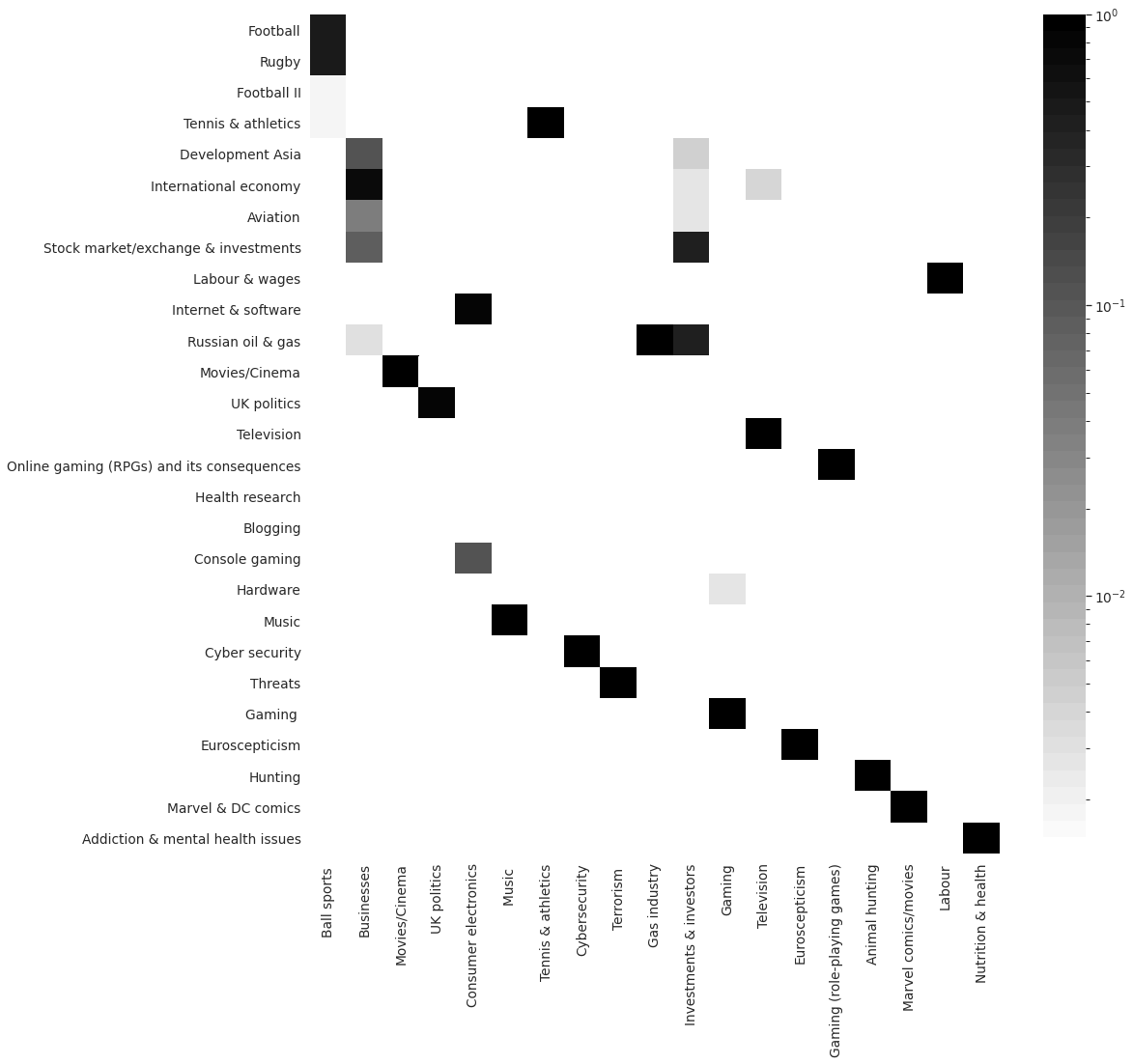}
	\caption{Heat map showing how topic terms of the 19 $\gamma=1.37$ topics (columns) get distributed on 27 topics at $\gamma=1.5$ (rows).}
	\label{fig:hm137to150}
\end{figure}

Continuing the route to $\gamma=2.0$ and $\gamma=2.5$, further specific topics can be detected: the separation of Tennis and Athletics, various country specific topics, additional industry sectors like the Automotive industry (see Table \ref{tab:autostratified}). All in all, this confirms the role of $\gamma$ as resolution parameter. However, as topics that are not easy to interpret become more frequent (see Table \ref{tab:varygamma}), one cannot reach arbitrarily high resolution.

\subsection{The Effect of Term Reduction}

In all of the above examples, we have used $G_{50}(\mathcal{D_{\rm BBC}})$, i.e., the BBC corpus with $p=50\%$ of its ranked terms. Here, we briefly investigate what happens when we work without reduction ($p=100\%$), or if we reduce much more ($p=10\%$).

With reference to Table \ref{tab:varypercentage}, we discuss the effect on topic number and topic interpretability. At low resolution ($\gamma=1.0$), we observe that increasing reduction (decreasing $p$) gives rise to a larger topic number. This is easily understandable as in the term community approach common terms are the glue which binds the community. Having more but less specific terms supports the formation of bigger clusters. On the other hand, smaller numbers of terms make big clusters unlikely and rather produce larger numbers of \mbox{smaller topics.}

\begin{specialtable}[H]
	\caption{Beginning of the stratified topic sheet for one of the 89 topics at $\gamma=2.5$ which can be interpreted 
	as Automotive. The terms in this clipping are dominated by car makes (LVMH being an outlier, though one can 
	imagine why it is included in the same cluster). While lower parts of the sheet not visible here also contain general 
	terms from the subject field, such as ``motorsport'', ``sportscar'', ``racing'', ``braking'', ``steering'',
	``throttle'', and ``gearbox'', 
	the specificity ranking puts the makes into higher position.}\label{tab:autostratified}
	\setlength{\cellWidtha}{\columnwidth/4-2\tabcolsep+0.0in}
\setlength{\cellWidthb}{\columnwidth/4-2\tabcolsep+0.0in}
\setlength{\cellWidthc}{\columnwidth/4-2\tabcolsep+0.0in}
\setlength{\cellWidthd}{\columnwidth/4-2\tabcolsep+0.0in}
\scalebox{1}[1]{\begin{tabularx}{\columnwidth}{
>{\PreserveBackslash\centering}m{\cellWidtha}
>{\PreserveBackslash\centering}m{\cellWidthb}
>{\PreserveBackslash\centering}m{\cellWidthc}
>{\PreserveBackslash\centering}m{\cellWidthd}}

\toprule

	      {Fiat} &            &            &            \\
	\hline
	general\_motor & Aston\_Martin & mitsubishi\_motor & Budget\_Aston \\
	\hline
	      grid &            &            &            \\
	\hline
	     Rover &            &            &            \\
	\hline
	      LVMH &            &            &            \\
	\hline
	   Peugeot &            &            &            \\
	\hline
	      Mini &            &            &            \\
	\hline
	   Ferrari &   Mercedes &   Maserati &            \\
	\hline
	   factory & Sindelfingen & Dingolfing & Ruesselsheim \\
	\hline
	Mitsubishi &   Chrysler & DaimlerChrysler &    vehicle \\
	\hline
	 Cadillacs &   Cadillac &  limousine &            \\
	\hline
	      saab &            &            &            \\
	\bottomrule

\end{tabularx}} 
\end{specialtable}

\vspace{-9pt}

\begin{specialtable}[H]
	\caption{Evaluation of topics produced from $G_{p}(\mathcal{D_{\rm BBC}})$ with resolution parameters 
	             $\gamma=1.0$ and $\gamma=2.0$ at three different values of the reduction percentage $p$. 
	             The table shows for each case the number $k$ of topics produced. The numbers $k^+$ and $k^-$ indicate 
	             how many topics the evaluators found easy to interpret (score 5 or 4) and hard to interpret (score 1 or 2){,} %AH: removed second comma
	             respectively; $\bar{z}$ is the mean score of the topics.}	\label{tab:varypercentage}
\setlength{\cellWidtha}{\columnwidth/6-2\tabcolsep+0.0in}
\setlength{\cellWidthb}{\columnwidth/6-2\tabcolsep-0.1in}
\setlength{\cellWidthc}{\columnwidth/6-2\tabcolsep+0.4in}
\setlength{\cellWidthd}{\columnwidth/6-2\tabcolsep-0.1in}
\setlength{\cellWidthe}{\columnwidth/6-2\tabcolsep-0.1in}
\setlength{\cellWidthf}{\columnwidth/6-2\tabcolsep-0.1in}
\scalebox{1}[1]{\begin{tabularx}{\columnwidth}{
>{\PreserveBackslash\centering}m{\cellWidtha}
>{\PreserveBackslash\centering}m{\cellWidthb}
>{\PreserveBackslash\centering}m{\cellWidthc}
>{\PreserveBackslash\centering}m{\cellWidthd}
>{\PreserveBackslash\centering}m{\cellWidthe}
>{\PreserveBackslash\centering}m{\cellWidthf}}

\toprule
	  \boldmath{$p$} & \boldmath{$\gamma$} &   \boldmath{$k$} &  \boldmath{$k^+$} &  \boldmath{$k^-$} &    \boldmath{$\bar{z}$} \\
	\midrule
	 100 &  1.0 &    5 &   5 &   0 &  5.0 \\
	 100 &  2.0 &   98 &  48 &  33 &  3.2 \\
	  50 &  1.0 &    8 &   8 &   0 &  4.8 \\
	  50 &  2.0 &   58 &  44 &   6 &  4.2 \\
	  10 &  1.0 &   38 &  18 &  11 &  3.1 \\
	  10 &  2.0 &  100 &  41 &  39 &  3.0 \\
	\bottomrule

\end{tabularx}} 
\end{specialtable}

While the lack of terms clearly is problematic for the topic interpretability at $\gamma=1.0$ and $p=10$, the case $\gamma=1.0$ and $p=100$ works well as it reproduces the 5-class structure of the corpus again\textemdash like for $\gamma=0.8$ and $p=50$. In fact, if the only purpose of topic detection was classification with respect to the coarse 5 classes of the corpus, the choice $\gamma=1.0$ and $p=100$ would be preferable, as a comparison of Tables \ref{tab:cross100} and \ref{tab:cstat100} \mbox{with Tables \ref{tab:cross} and \ref{tab:cstat}}, respectively,  shows. However, the unreduced $p=100$ wins its slightly better ability to predict the dominant topic at the cost of recognizing secondary topics less well.

\begin{specialtable}[H]
	\caption{Crosstable between preassigned classes and detected dominant topics ($p=100$, $\gamma=1.0$) for all documents of the BBC corpus.}\label{tab:cross100}
\setlength{\cellWidtha}{\columnwidth/6-2\tabcolsep+0.3in}
\setlength{\cellWidthb}{\columnwidth/6-2\tabcolsep-0.1in}
\setlength{\cellWidthc}{\columnwidth/6-2\tabcolsep+0.1in}
\setlength{\cellWidthd}{\columnwidth/6-2\tabcolsep-0.1in}
\setlength{\cellWidthe}{\columnwidth/6-2\tabcolsep-0.1in}
\setlength{\cellWidthf}{\columnwidth/6-2\tabcolsep-0.1in}
\scalebox{1}[1]{\begin{tabularx}{\columnwidth}{
>{\PreserveBackslash\centering}m{\cellWidtha}
>{\PreserveBackslash\centering}m{\cellWidthb}
>{\PreserveBackslash\centering}m{\cellWidthc}
>{\PreserveBackslash\centering}m{\cellWidthd}
>{\PreserveBackslash\centering}m{\cellWidthe}
>{\PreserveBackslash\centering}m{\cellWidthf}}

\toprule
	\textbf{Topic Class} &  \textbf{Economy} &  \textbf{Entertainment} &  \textbf{Politics} &  \textbf{Sports} &  \textbf{Technology} \\
	  
	\midrule
	Business      &      491 &              1 &        12 &       0 &           6 \\
	Entertainment &        8 &            354 &        12 &       0 &          12 \\
	Politics      &       12 &              0 &       402 &       2 &           1 \\
	Sport         &        2 &              0 &         2 &     507 &           0 \\
	Tech          &       12 &              7 &         6 &       6 &         370 \\
	\bottomrule

\end{tabularx}} 
\end{specialtable}

\begin{specialtable}[H]
	\caption{Classification statistics for predicting preassigned classes by detected dominant topics ($p=100$, $\gamma=1.0$); to be compared with Table \ref{tab:cstat}.
	Note that {Table} %MDPI: Wrong order. Tables should be mentioned in numerical order. %AH: This is a deliberate lookahead, pointing to another summarizing table in a later section. This information will be useful to the reader who wants to compare results of various methods.
12 contains a combined view of f1-scores for all classification statistics for easier comparison.}\label{tab:cstat100}
	\setlength{\cellWidtha}{\columnwidth/4-2\tabcolsep+0.0in}
\setlength{\cellWidthb}{\columnwidth/4-2\tabcolsep+0.0in}
\setlength{\cellWidthc}{\columnwidth/4-2\tabcolsep+0.0in}
\setlength{\cellWidthd}{\columnwidth/4-2\tabcolsep+0.0in}
\scalebox{1}[1]{\begin{tabularx}{\columnwidth}{
>{\PreserveBackslash\centering}m{\cellWidtha}
>{\PreserveBackslash\centering}m{\cellWidthb}
>{\PreserveBackslash\centering}m{\cellWidthc}
>{\PreserveBackslash\centering}m{\cellWidthd}}
	\toprule
	{} &  \textbf{Precision} &    \textbf{Recall} &  \textbf{f1-Score} \\
	\midrule
	Business      &   0.935 &  0.963 &  0.949 \\
	Entertainment &   0.978 &  0.917 &  0.947 \\
	Politics      &   0.926 &  0.964 &  0.945 \\
	Sport         &   0.984 &  0.992 &  0.988 \\
	Tech          &   0.951 &  0.923 &  0.937 \\
	\midrule
	weighted avg  &   0.955 &  0.955 &  0.955 \\
	\bottomrule

\end{tabularx}} 
\end{specialtable}

At higher resolution, $\gamma=2.0$, the advantages of a moderate reduction of topic terms are clearly visible: $p=100$ and $p=10$ produce a comparable number of well interpretable topics as $p=50$, but both extreme choices also yield many uninterpretable topics. These typically consist of few topic terms. In the case of $p=100$, many unspecific terms are involved, whereas in the case of $p=10$ the topic terms are so narrow that they seem to belong to one document only.

Term reduction is about finding the right balance between removing as many unspecific terms as possible and keeping enough terms for characterizing topics in detail. While there is no practicable way to predict the optimal value of $p$, we can provide some guidelines, based on the results presented here as well as on findings from further tests and work with different text corpora. These include summaries of scientific studies~\cite{Hamm2020} and parliamentary documents~\cite{Odrowski2020}, but also abstracts of scientific articles, RSS news feeds, and mixed corpora. More generally, the longer the documents are, the stronger the reduction (i.e., the lower $p$) should be. More specifically, for articles with one or two pages, reductions between 50\% and 25\% work\textemdash by and large equally\textemdash well.  Short texts like abstracts work better with less reduction, whereas for long documents with many pages reduction between 10\% and 20\% is helpful, also as it decreases the term network size and consequently the computational effort.

\section{Topic Interpretability and Comparison with LDA}
\label{sec:interpretability_comparison}

The term community method for topic detection described above is geared towards good topic interpretability. The results presented in Section \ref{sec:influence} confirm that evaluators 
found the topics uncovered by the method to be of high quality in that sense. Before we look in more detail into the factors that determine the topic interpretability we apply standard LDA to the same corpus for the sake of comparison.

\subsection{LDA Topics for the BBC Corpus}

LDA is based on the assumption of a generative process where topics are considered to be probability distributions over all words of the corpus, which are not directly observable but latent in the documents in that each document is a random mixture of topics. The word distributions within the topics as well as the topic distributions within the documents are assumed to have been drawn from a Dirichlet distribution. Fixing the number $k$ of topics within the corpus and further hyper parameters that determine the shape of the Dirichlet distribution, methods of statistical inference can be used to determine the latent distributions of words per topic from the observed distributions of words per document. In particular, the method of Gibbs sampling is known to produce convincing topics in \mbox{many applications}.

This motivates a comparison of the topics detected as term communities with the topics identified using LDA. For generating these topics, we used the popular Mallet LDA toolkit~\cite{McCallum2002} through the Python wrapper contained in the library Gensim~\cite{Rehurek2011}. We fixed the topic number $k$ to values described below and used the defaults for all other parameters. This means that further tuning might improve the results shown below, but also that we compare with the typical way in which LDA is used in the applied literature.

The resulting topics were presented to the evaluators as lists of terms, sorted by their probability, where only terms with a probability greater than 0.001 were shown. The labeling and evaluation process was carried out in the same way as for the term communities. Table \ref{tab:varylda} shows how the evaluators graded the topics for 5 different values of $k$. For reasons of comparison we also include rows from Table \ref{tab:varygamma} for term communities that resulted in the same number of topics. Altogether, the table shows that the evaluators found most LDA topics well interpretable, but the scores are consistently below the results for the term community method. 

\begin{specialtable}[H]
	\caption{Evaluation of topics of the BBC corpus produced with a standard LDA procedure in comparison to term community
	              topics in $G_{50}(\mathcal{D_{\rm BBC}})$.
	              In each block, the table shows an LDA model and the corresponding term communities (TeCo)  
	              resulting in the same number of topics and lists for each model 
	              the number $k$ of topics produced, the numbers $k^+$ and $k^-$ indicating how many topics the 
	              evaluators found easy to interpret (score 5 or 4) and hard to interpret (score 1 or 2), respectively, 
	              and the mean score $\bar{z}$ of the topics.
	              For LDA models the coherence value $c_v$ is also given.}\label{tab:varylda}
\setlength{\cellWidtha}{\columnwidth/6-2\tabcolsep+0.5in}
\setlength{\cellWidthb}{\columnwidth/6-2\tabcolsep-0.1in}
\setlength{\cellWidthc}{\columnwidth/6-2\tabcolsep-0.1in}
\setlength{\cellWidthd}{\columnwidth/6-2\tabcolsep-0.1in}
\setlength{\cellWidthe}{\columnwidth/6-2\tabcolsep-0.1in}
\setlength{\cellWidthf}{\columnwidth/6-2\tabcolsep-0.1in}
\scalebox{1}[1]{\begin{tabularx}{\columnwidth}{
>{\PreserveBackslash\centering}m{\cellWidtha}
>{\PreserveBackslash\centering}m{\cellWidthb}
>{\PreserveBackslash\centering}m{\cellWidthc}
>{\PreserveBackslash\centering}m{\cellWidthd}
>{\PreserveBackslash\centering}m{\cellWidthe}
>{\PreserveBackslash\centering}m{\cellWidthf}}

	\toprule
	 \textbf{Model} &   \boldmath{$k$} &  \boldmath{$k^+$} &  \boldmath{$k^-$} &   \boldmath{$\bar{z}$} & \boldmath{$c_v$} \\
	\midrule
	 LDA $k=5$ &   5 &   5 &   0 &  5.0 & 0.59 \\
	 TeCo $\gamma=0.8$ &   5 &   5 &   0 &  5.0 & \\
	\midrule
	 LDA $k=10$ &  10 &   9 &   0 &  4.4 & 0.62 \\
	 TeCo $\gamma=1.07$ &  10 &  10 &   0 &  4.8 & \\
	\midrule
	 LDA $k=19$ &  19 &  13 &   1 &  4.1 & 0.61 \\
	 TeCo $\gamma=1.37$ &  19 &  18 &   0 &  4.7 & \\
	\midrule
	 LDA $k=27$ &  27 &  20 &   4 &  4.0 & 0.57 \\
	 TeCo $\gamma=1.5$ &  27 &  25 &   1 &  4.5 & \\
	\midrule
	 LDA $k=58$  &  58 &  39 &   9 &  3.7 & 0.57 \\
	 TeCo $\gamma=2.0$ &  58 &  44 &   6 &  4.2 & \\
\bottomrule

\end{tabularx}} 
\end{specialtable}

The evaluators report that cognitive processing of stratified term clusters (with a significant proportion of named entities), as they were presented in the case of term communities, appears prima facie more complex than that of word lists (with more general terms), as in LDA. After familiarizing with both ways of presentation through interpreting a couple of topics, however, more information at a glance and more details eventually increase the interpretability of topics. In particular, this results in fewer unidentifiable topics and better differentiability between topics. Furthermore, while interpreting topics from both methods becomes more difficult with increasing topic numbers, this effect is stronger for LDA.

Conversely, stratified word clusters including a decent amount of named entities were considered very fruitful from a domain application perspective: generally speaking, more, and more detailed, information (actors, issues, places, time references, aspects) about the topics and the corpus as such is beneficial for most social science purposes, and hardly ever an impediment.

The clearest results are obtained at $k=5$, corresponding to the number of preassigned classes. This model can be evaluated as a classifier in the same way we did for term community topics at $p=50, \gamma=0.8$ and at $p=100, \gamma=1.0$. The results are presented \mbox{in Tables \ref{tab:crosslda} and \ref{tab:cstatlda}}. Precision and recall are high, but clearly below the values for the term community method.

\begin{specialtable}[H]
	\caption{Crosstable between preassigned classes and detected dominant topics according to LDA with $k=5$ for all documents of the BBC corpus; to be compared with 
	               Tables \ref{tab:cross} and \ref{tab:cross100}.}\label{tab:crosslda}
\setlength{\cellWidtha}{\columnwidth/6-2\tabcolsep+0.3in}
\setlength{\cellWidthb}{\columnwidth/6-2\tabcolsep-0.1in}
\setlength{\cellWidthc}{\columnwidth/6-2\tabcolsep+0.1in}
\setlength{\cellWidthd}{\columnwidth/6-2\tabcolsep-0.1in}
\setlength{\cellWidthe}{\columnwidth/6-2\tabcolsep-0.1in}
\setlength{\cellWidthf}{\columnwidth/6-2\tabcolsep-0.1in}
\scalebox{1}[1]{\begin{tabularx}{\columnwidth}{
>{\PreserveBackslash\centering}m{\cellWidtha}
>{\PreserveBackslash\centering}m{\cellWidthb}
>{\PreserveBackslash\centering}m{\cellWidthc}
>{\PreserveBackslash\centering}m{\cellWidthd}
>{\PreserveBackslash\centering}m{\cellWidthe}
>{\PreserveBackslash\centering}m{\cellWidthf}}
	\toprule
	\textbf{Topic Class} &  \textbf{Economy} &  \textbf{Music} \& \textbf{Films} &  \textbf{Politics} &  \textbf{Sports} &  \textbf{Technology} \\
	
	\midrule
	Business      &      482 &              4 &        14 &       0 &          10 \\
	Entertainment &       14 &            355 &         7 &       1 &           9 \\
	Politics      &       40 &              8 &       358 &       4 &           7 \\
	Sport         &       15 &             38 &        14 &     444 &           0 \\
	Tech          &       15 &             12 &         2 &      24 &         348 \\
\bottomrule

\end{tabularx}} 
\end{specialtable}

\vspace{-9pt}

\begin{specialtable}[H]
	\caption{Classification statistics for predicting preassigned classes by detected dominant topics from LDA with $k=5$; to be compared with Tables \ref{tab:cstat} 
	               and \ref{tab:cstat100}. For easier direct comparison, we repeat the f1-scores of those Tables here in the rightmost two columns.}	\label{tab:cstatlda}
\setlength{\cellWidtha}{\columnwidth/6-2\tabcolsep+0.2in}
\setlength{\cellWidthb}{\columnwidth/6-2\tabcolsep-0.2in}
\setlength{\cellWidthc}{\columnwidth/6-2\tabcolsep-0.2in}
\setlength{\cellWidthd}{\columnwidth/6-2\tabcolsep-0.2in}
\setlength{\cellWidthe}{\columnwidth/6-2\tabcolsep+0.2in}
\setlength{\cellWidthf}{\columnwidth/6-2\tabcolsep+0.2in}
\scalebox{1}[1]{\begin{tabularx}{\columnwidth}{
>{\PreserveBackslash\centering}m{\cellWidtha}
>{\PreserveBackslash\centering}m{\cellWidthb}
>{\PreserveBackslash\centering}m{\cellWidthc}
>{\PreserveBackslash\centering}m{\cellWidthd}
>{\PreserveBackslash\centering}m{\cellWidthe}
>{\PreserveBackslash\centering}m{\cellWidthf}}
	\toprule
	{} & &  \textbf{LDA} & & \textbf{TeCo} & \textbf{TeCo}\\
	{} & & \boldmath{$k=5$} & & \boldmath{$p=50, \gamma=0.8$} &  \boldmath{$p=100, \gamma=1.0$}\\
	{} &  \textbf{Precision} &    \textbf{Recall} &  \textbf{f1-Score}&  \textbf{f1-Score}  &  \textbf{f1-Score}\\
	\midrule
	Business &   0.852 &  0.945 &  0.896 &  0.933 &  {0.949} %MDPI: Is the bold necessary? If yes, please add explanation for it. If not, please cancel. %AH: It is actually not very useful here, so it does not need to be bold.
 \\
	Entertainment &   0.851 &  0.920 &  0.884 &   0.933 &  0.947 \\
	Politics &   0.906 &  0.859 &  0.882 &  0.926 &  0.945 \\
	Sport &   0.939 &  0.869 &  0.902  &  0.987 &  0.988 \\
	Tech  &   0.930 &  0.868 &  0.898 &  0.931 &  0.937 \\
	\midrule
	weighted avg   &   0.896 &  0.893 &  0.893  &  0.944  &  0.955 \\
\bottomrule

\end{tabularx}} 
\end{specialtable}

In fact, with typical LDA reasoning one would argue that one should not work with $k=5$ at all but rather tune $k$ such as to achieve an optimal result. Very often optimization of LDA hyperparameters is understood to be targeted at maximizing coherence of the topics. According to~\cite{Roeder2015}, one of several coherence measures, usually called $c_v$, is especially well correlated with human evaluation of coherence. It is calculated from the co-occurrence statistics of topic words within text windows in the corpus documents. Table \ref{tab:varylda} shows this value for the various LDA models. In the present situation, $c_v$ reaches its maximum at $k=10$ with another local maximum at $k=19$; $k=5$ and the two higher $k$ values are clearly suboptimal with respect to $c_v$. Our evaluators' scores are not convincingly correlated to $c_v$. 

It is tempting to draw from the results presented so far the conclusion that LDA and term communities offer two options for finding the same topics, both normally working well, with slight advantages concerning interpretability on the side of the term communities. Instead, further analysis of the term composition of the topics shows that both methods find different topics. This can be seen in the heat map in Figure \ref{fig:hmldatoteco}. 

The columns show the LDA topics found at $k=10$, the rows display the term communities for $P=50, \gamma=1.07$. The heat map grayscale indicates how many of the most probable terms of an LDA topic come from which term community. If there was a 1-to-1 correspondence of topics, the heat map would only show exactly one single dark square for every row and every column. However, this is not the case. Rather, there are LDA topics that extend over several term communities (e.g., Entertainment comprising Music and Movies), and there are term communities that contain several LDA topics (e.g., Technology \& (video) gaming containing Consumer electronics and Computer \& Internet). In LDA, Sports is a very broad topic, Financial market is a very narrow topic. In term communities, Economy and UK politics are very broad, while the communities for TV and for (Video) Gaming are so small that they do not contain any of the most likely LDA terms.

Altogether, LDA and term communities seem to offer complementary views of the subjects discussed in the corpus.

\begin{figure}[H]
	%\centering
	\includegraphics[width=380bp]{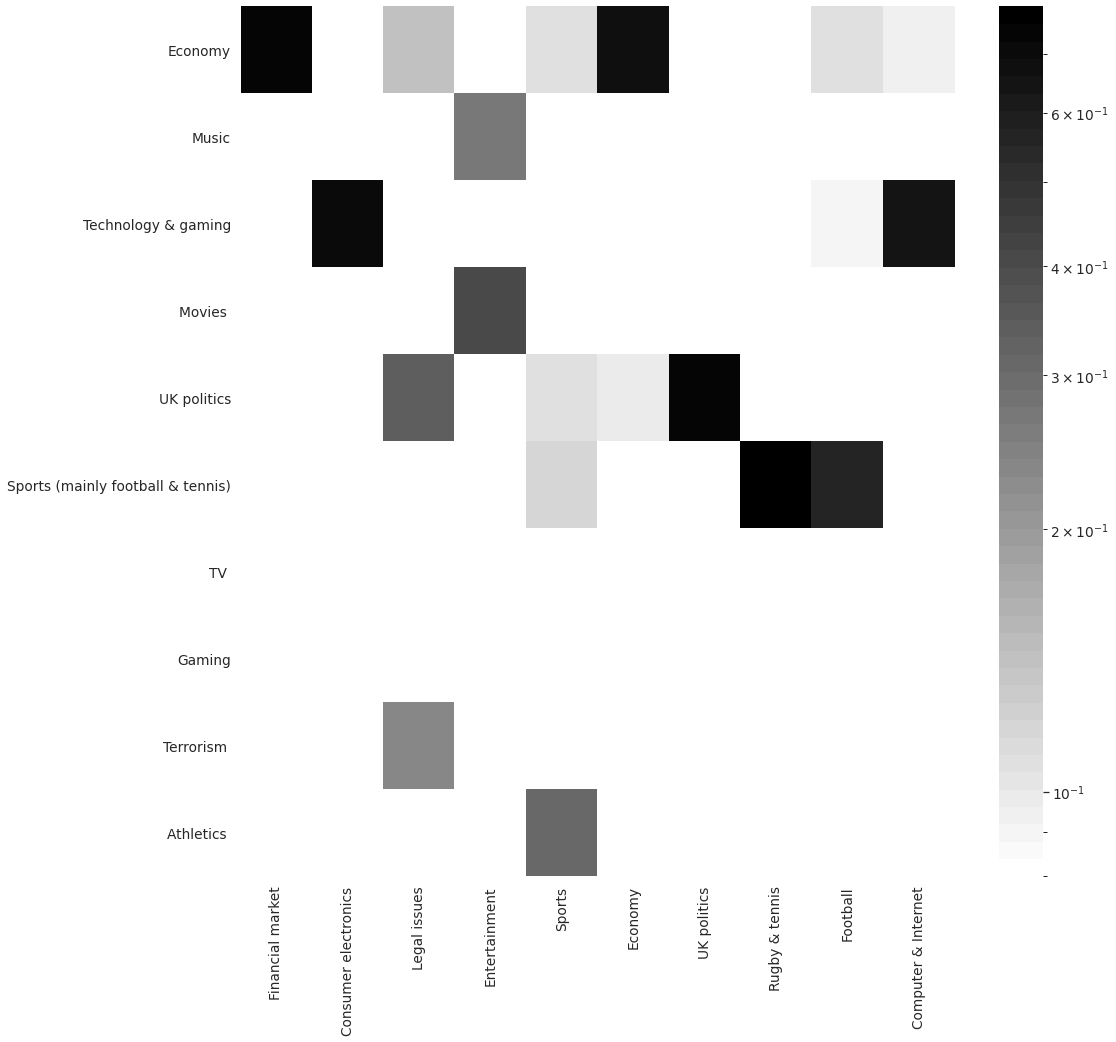}
	\caption{Heat map showing how topic terms of the $k=10$ LDA topics (columns) get distributed on 10 term community topics at $\gamma=1.07$ (rows).}
	\label{fig:hmldatoteco}
\end{figure}

\subsection{Factors Influencing the Topic Interpretability}

Based on the evaluation of the interpretability of many topics produced with two totally different methods of topic evaluation, we want to look into factors which make the difference between topics that can be recognized easily and topics that defy interpretation. Therefore, we asked the evaluators for each topic with low score what caused the difficulties with interpretation. One obvious reason which affected both\textemdash topics presented as term communities as well as LDA topic term distributions\textemdash was that topic terms seemed to point into contradictory directions. However, there were two other equally important reasons: first, some topics consisted of only a few dozen terms without projecting a clear picture about a common theme\textemdash this was a relatively frequent problem for the term community method; second, there were topics that showed almost exclusively generic terms, which lacked expressiveness and made it impossible to recognize a specific theme\textemdash this is a common problem for LDA. These two effects can be traced back to a common root: the lack of informative terms. To put it differently: A topic does not only need non-contradictory but also informative terms in order to be easily interpretable.

The corpus term ranking which we established in Section \ref{sec:method} offers a way to separate informative from non-informative terms\textemdash also when talking about terms in the LDA term distributions: we consider a term only as informative if its ranking value is higher than a certain threshold. For the BBC corpus it works well to define the set of highly informative topic terms of topic $C$ as $H=\{t\in C: r(t)>0.25\}$. In order to assess the degree of informative terms in a topic, we simply count $M_H = \# H$.

Assessing the risk of contradictions within the highly informative terms can be done by some form of coherence measure. Rather than using a self-referential intrinsic measure that compares to co-occurrences within the corpus documents or working with a larger external corpus, we suggest to use a word embedding coherence as defined in~\cite{Fang2016} which we base here on the same fastText embedding that we have used already in Section \ref{sec:method}, as it derives relatedness of terms from an extremely large text collection:
\[
 c_H^{\rm emb} = \frac{1}{M_H(M_H-1)} \sum_{s,t \in H \atop s\neq t}  \fTsim(s,t)
\]
where $\fTsim(s,t)$ is the cosine similarity between the two fastText vectors for $s$ and $t$.

Figure \ref{fig:interpretability} shows how topics which were evaluated as difficult for interpretation are positioned with respect to the two measures $M_H$ (amount of informative terms) and $c_H^{\rm emb}$ (consistency of informative terms). Included are 85 topics from term community detection ($P=50$ with $\gamma=1.37$ and $\gamma=1.5$) and 85 topics from LDA ($k=27$ and $k=58$). Topics with good interpretability are colored gray, the 20 topics for which the evaluators gave low scores are colored black. Term community topics are marked as circles, LDA topics as diamonds. Obviously, all problematic topics are gathered in the lower left quadrant of relatively small coherence and small term number, confirming the above hypothesis about the two factors that can cause difficulties for interpretation.

\begin{figure}[H]
	%\centering
	\includegraphics[width=350bp]{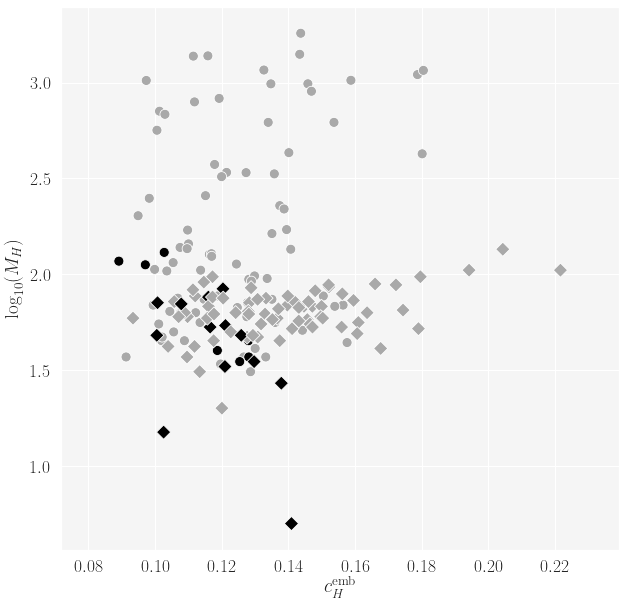}
	\caption{Scatterplot visualizing a relation between topic interpretability, term specificity, and coherence. LDA topics are represented through diamonds, term community topics
	               through circles. Topics that the evaluators scored with grade 3 or higher are colored gray; black color marks the topics that were found difficult to interpret.
	               The topics are positioned vertically according to the logarithm  of the number of its highly ranked terms $M_H$ and horizontally according to 
	               the embedding coherence $c_H^{\rm emb}$ of the highly ranked terms.}  
	\label{fig:interpretability}
\end{figure}

This also shows that the coherence measure alone is not a criterion which can predict topic interpretability, at least not when comparing different methods of topic detection. In fact, even a combination of coherence and term number cannot tell whether or not evaluators will find it easy to recognize the meaning of a topic. There are several examples in the left lower quadrant where topics were clear. However, producing topics with many informative terms and considerable word embedding coherence does reduce the risk of ending up with meaningless topics.

\section{Conclusions and Outlook}
\label{seq:conclusions}

Term community detection using parametrized modularity in a rank-reduced term co-occurrence network results in topics that are nearly always easy to interpret by domain experts. This observation is fully substantiated by the extensive studies on the BBC news corpus presented here. The ability to produce topics on different resolution levels by varying a continuous parameter is a feature that is particularly relevant from a domain expert perspective since text corpora are assumed to contain topics on several levels of granularity.  For example, one can focus on broader thematic areas, particular discourses, or specific issues.
Our special form of term ranking on corpus level and additional clustering in word embedding space has proven to be essential for in-depth topic interpretation derived from term communities.

Term community detection is one of many methods for discovering topics in large text collections. While it is natural to ask which of those methods works best, one can hardly expect a clear answer to this question: one reason is that no quantitative criterion is known that adequately predicts the human interpretability of a topic; as we have seen, topic coherence alone is certainly not sufficient. Another reason derives from the fact that certain methods are known to work better or worse depending on some properties of the corpus\textemdash a one-suits-all method might not exist. This is particularly true for generative topic models which always are based on quite specific and hardly verifiable assumptions about modeling details like the conditional dependence structure and prior distributions.

The phenomenological approach of term community detection however has worked convincingly well for the example corpus studied here and can be recommended for initial overview but also for deeper insight with higher resolution when investigating unknown corpora. Nevertheless, we saw when comparing with LDA topics that it certainly is interesting to run different methods on the same corpus as this may produce \mbox{complementary topics.}

The fact that different topic detection methods find different topics can be compared to a recent observation for community detection~\cite{Riolo2020}, where typically the communities also show considerable variation depending on the method used. There it was pointed out that all the differing communities can still be seen as different arrangements of the same building blocks. The same might be true for topics: they may be conceived of as arrangements (of elementary events, incidents, and concepts) that emerge with the corpus itself, rather than as pre-existing ideas from which the authors of documents could choose at the time of writing. The various topic detection methods search for such arrangements in different ways, thereby shedding light on the discourse underlying the corpus from distinct, but most likely complementary, perspectives.

In the regular work of our group, the method presented here has turned out to be straightforward and productive for monitoring diverse corpora in the course of strategic analyses: political documents, scientific publications, government notifications, research news, general news. For this purpose, we have developed TeCoMiner~\cite{Hamm2020}, a software tool for interactively investigating topics as term communities. There we take advantage of the fact that computing time for finding term communities is relatively small, faster than running a sufficient number of Gibbs sampling iterations for LDA. Furthermore, the stratified view of topic terms lends itself to very comprehensive but lucid visualizations.

Regarding future work, it will certainly be worthwhile amalgamating the network-theoretical approach to topic detection with other network-theoretical text mining procedures like co-authorship or citation networks, and with knowledge graphs.

The method presented here is likely to be particularly well suited for applications in the social sciences, especially as it produces both informative and well interpretable topics on different levels of thematic granularity. It should therefore be considered as a promising alternative to probabilistic topic modeling.

\vspace{6pt}
%%%%%%%%%%%%%%%%%%%%%%%%%%%%%%%%%%%%%%%%%%%%%%%
%%%%%%%%%%%%%%%%%%%%%%%%%%%%%%%%%%%%%%%%%%%%%%%
%%%%%%%%%%%%%%%%%%%%%%%%%%%%%%%%%%%%%%%%%%%%%%%

%%%%%%%%%%%%%%%%%%%%%%%%%%%%%%%%%%%%%%%%%%
%% optional
\supplementary{{The following} %MDPI: please cite Supplementary in main text. %AH: Now cited in the introduction of Section 4
 are available online at \linksupplementary{s1}: Excel files (.xlsx) with full details of all topics within the example corpus detected by the various methods described in the text; see the included readme.txt.
}


%%%%%%%%%%%%%%%%%%%%%%%%%%%%%%%%%%%%%%%%%%
\authorcontributions{Conceptualization, A.H. and S.O.; methodology, A.H.; software, A.H.; validation, S.O. and A.H.; investigation, S.O.; writing---original draft preparation, A.H.; writing---review and editing, S.O. All authors have read and agreed to the published version of the manuscript.}

\funding{This research received no external funding.}

\dataavailability{The corpus of BBC news articles which was used for this study was originally compiled by D. Greene and P. Cunningham for~\cite{Greene2006} and can be downloaded from \url{http://mlg.ucd.ie/datasets/bbc.html} (accessed on 20 May 2021)%MDPI: Please add accessed date.
.
Complete sheets of topic terms and derived data obtained with the methods described in this article can be found in the \mbox{Supplementary Materials.} 
} 

\acknowledgments{We are deeply grateful to Jana Thelen, whose work on a predecessor version of the method described here motivated and shaped several of its details. Special thanks go to Rasmus Beckmann and Mark Azzam for fruitful discussions and constant support. We also acknowledge the domain expert support of Friderike Uphoff and Stefan Odrowski in evaluating the topics.}

\conflictsofinterest{The authors declare no conflict of interest.} 

%%%%%%%%%%%%%%%%%%%%%%%%%%%%%%%%%%%%%%%%%%
%% Only for journal Encyclopedia
%\entrylink{The Link to this entry published on the encyclopedia platform.}

%%%%%%%%%%%%%%%%%%%%%%%%%%%%%%%%%%%%%%%%%%
%% Optional
\abbreviations{Abbreviations}{The following abbreviations are used in this manuscript:\\

\noindent 
\begin{tabular}{@{}ll}
LDA & Latent Dirichlet Allocation\\
TeCo& Term Communities\\
LSI & Latent Semantic Indexing\\
pLSA & probabilistic Latent Semantic Analysis\\
NMF & Non-negative Matrix Factorization\\
STM & Structural Topic Models\\
NLP & Natural Language Processing\\
CBOW & Continuous Bag Of Words\\
RSS & Really Simple Syndication\\
BBC & British Broadcasting Corporation 
\end{tabular}}

%%%%%%%%%%%%%%%%%%%%%%%%%%%%%%%%%%%%%%%%%%

\end{paracol}
\reftitle{References}

\end{document}